\documentclass[twoside]{article}

\usepackage{multirow}
\usepackage{rotating}
\usepackage{booktabs}
\usepackage{makecell}
\usepackage{amssymb}
\usepackage{bbding}

\usepackage[accepted]{aistats2023}

\usepackage[round]{natbib}

\begin{document}

\runningtitle{Weather2K: A Multivariate Spatio-Temporal Benchmark Dataset for Meteorological Forecasting}

\twocolumn[

\aistatstitle{Weather2K: A Multivariate Spatio-Temporal Benchmark Dataset\\
for Meteorological Forecasting Based on Real-Time Observation Data\\ 
from Ground Weather Stations}

\aistatsauthor{Xun Zhu\textsuperscript{1, \dag}
\qquad \ \ Yutong Xiong\textsuperscript{1, \dag}
\qquad \ \ Ming Wu\textsuperscript{1, \ddag}
\qquad \ \ Gaozhen Nie\textsuperscript{2}
\qquad \ \ Bin Zhang\textsuperscript{1}
\qquad \ \ Ziheng Yang\textsuperscript{1} }

\aistatsaddress{$^1$Beijing University of Posts and Telecommunications, China
\\$^2$National Meteorological Center, China Meteorological Administration, China}]

\begin{abstract}
Weather forecasting is one of the cornerstones of meteorological work.
In this paper, we present a new benchmark dataset named Weather2K, which aims to make up for the deficiencies of existing weather forecasting datasets in terms of real-time, reliability, and diversity, as well as the key bottleneck of data quality.
To be specific, our Weather2K is featured from the following aspects: 
1) Reliable and real-time data.
The data is hourly collected from 2,130 ground weather stations covering an area of 6 million square kilometers.
2) Multivariate meteorological variables. 
20 meteorological factors and 3 constants for position information are provided with a length of 40,896 time steps.
3) Applicable to diverse tasks. We conduct a set of baseline tests on time series forecasting and spatio-temporal forecasting. 
To the best of our knowledge, our Weather2K is the first attempt to tackle weather forecasting task by taking full advantage of the strengths of observation data from ground weather stations.
Based on Weather2K, we further propose Meteorological Factors based Multi-Graph Convolution Network (MFMGCN), which can effectively construct the intrinsic correlation among geographic locations based on meteorological factors.
Sufficient experiments show that MFMGCN improves both the forecasting performance and temporal robustness.
We hope our Weather2K can significantly motivate researchers to develop efficient and accurate algorithms to advance the task of weather forecasting.
The dataset can be available at https://github.com/bycnfz/weather2k/.
\end{abstract}

\section{INTRODUCTION}

The changing weather has been profoundly affecting people’s lives since the beginning of mankind.
Weather conditions play a crucial role in production and economic industries such as transportation, tourism, agriculture and energy.
Therefore, reliable and efficient weather forecasting is of great economic, scientific and social significance.
The weather forecasting task deserves extensive attention.

Meteorological factors, such as temperature, humidity, visibility, and precipitation, can provide strong support and historical information for researchers to analyze the variation tendency of weather. 
For the past few decades, Numerical Weather Prediction (NWP) is the widely used traditional method, which utilizes physical models to simulate and predict meteorological dynamics in the atmosphere or on the Earth's surface \citep{muller2014massively}.
However, the prediction of NWP may not be accurate enough due to the uncertainty of the initial conditions of the differential equation \citep{tolstykh2005some}, especially in complex atmospheric processes. In addition, NWP has high requirements on computing power.

In recent years, meteorological researchers have achieved considerable breakthroughs and successes in introducing data-driven approaches, most prominently deep learning methods, to the task of weather forecasting.
Data-driven approaches exploit the historical meteorological observation data aggregated over years to model patterns to learn the input-output mapping. 
In the task of time series forecasting, Transformers have shown great modeling ability for long-term dependencies and interactions in sequential data benefiting from the self-attention mechanism.
In the more challenging task of spatio-temporal forecasting, the classical Convolutional Neural Networks (CNNs) which work well on regular grid data in Euclidean domain have been greatly challenged to handle this problem due to the irregular sampling of most spatio-temporal data.
The Graph Neural Networks (GNNs), which have already been extensively applied to traffic forecasting, yield effective and efficient performance by properly treating the variables and the connections among them as graph nodes and edges, respectively.
However, studies focusing on the field of meteorology are relatively scarce, while the demand for weather forecasting is increasing dramatically.

For data-centric deep learning method, the performance of the model heavily depends on the quality of the available training data.
High-quality benchmark datasets can serve as “catalysts” for quantitative comparison between different algorithms and promote constructive competition \citep{ebert2017vision}. 
In the field of meteorological science, reanalysis dataset cannot ensure the authenticity of the data.
Remote sensing dataset cannot reliably reflect complexities of near-surface weather conditions.
Fusion dataset cannot guarantee real-time performance.
Inadequacies of existing datasets also include diversity of meteorological factors and applicable tasks.
In this work, we present a new benchmark dataset named Weather2K, aiming at advancing the progress on weather forecasting tasks based on deep learning methods.
As far as we know, our Weather2K is the first attempt to tackle the challenging weather forecasting task by entirely using the observation data from the ground weather stations.
Concretely, our Weather2K has the following characteristics: 

\textbf{Reliable and Real-time Data}
The raw data is built on the one-hourly observation data of 2,130 ground weather stations from the China Meteorological Administration (CMA), which can be updated hourly in real-time. 
We have spent considerable time on data processing to ensure continuous temporal coverage and consistent data quality.

\textbf{Multivariate Meteorological Variables}
Based on meteorological consideration, 20 important near-surface meteorological factors and 3 time-invariant constants for position information are provided in the Weather2K.

\textbf{Applicable to Diverse Tasks}
We provide two versions of Weather2K for different application directions: time series forecasting and spatio-temporal forecasting.
Well-arranged spatio-temporal sequences can be easily selected and accessed for different tasks of weather forecasting.

As a benchmark, we conduct a set of baseline tests on two application directions to validate the performance of our Weather2K.
4 representative transformer-based models and 8 state-of-the-art spatio-temporal GNN models are set up for comparison in time series forecasting and spatio-temporal forecasting, respectively.
We further propose Meteorological Factors based Multi-Graph Convolution Network (MFMGCN).
Considering utilizing multivariate meteorological information, MFMGCN fuses 4 static graphs representing different types of spatio-temporal information and 1 dynamic graph to model correlations among geographic locations, and also uses a complete convolutional structure followed time-space consistency.
We validate both the improvement and temporal robustness of MFMGCN on Weather2K through extensive experiments.

We hope our Weather2K can significantly motivate more researchers to develop efficient and accurate algorithms and help ease future research in weather forecasting task.


\section{PREVIOUS WORK}

\subsection{Meteorological Science Datasets}

There are many types and sources of data commonly used in the meteorological science community.
Tropical Rainfall Measuring Mission (TRMM) precipitation rate dataset \citep{kummerow2000status} and the Integrated Multi-satellite Retrievals for GPM (IMERG) \citep{huffman2015nasa} are developed on remote sensing data from satellites.
Multi-Source Weighted-Ensemble Precipitation (MSWEP) \citep{beck2017mswep} merges satellite and reanalysis data.
The ERA5 \citep{hersbach2020era5} is the most advanced global reanalysis product created by the European Center for Medium Weather Forecasting (ECMWF).
The Global Land Data Assimilation System (GLDAS) \citep{rodell2004global} and the Canadian Land Data Assimilation System (CaLDAS) \citep{carrera2015canadian} are well-known gridded datasets at global and regional scales, respectively.
It is worth noting that the China Meteorological Forcing Dataset (CMFD) \citep{he2020first} is the first gridded near-surface meteorological dataset with high spatio-temporal resolution.
Differently, our Weather2K is the first truly meaningful meteorological science dataset entirely based on the observation data from ground weather stations in China.

\subsection{Weather Forecasting Datasets}
In the task of weather forecasting, different datasets are constructed in various ways. And the number of involved meteorological factors varies widely.
ExtremeWeather \citep{racah2017extremeweather} is an image dataset for detection, localization, and understanding of extreme weather events.
CloudCast \citep{nielsen2021cloudcast} is a satellite-based image dataset for forecasting 10 different cloud types.
Jena Climate (https://www.bgc-jena.mpg.de/wetter/) is made up of 14 meteorological factors recorded over several years at the weather station of the Max Planck Institute for Biogeochemistry.
Climate Change (http://berkeleyearth.org/data/) provided by the Berkeley Earth focuses on global land and ocean temperature data.
WeatherBench \citep{rasp2020weatherbench} is a benchmark dataset for data-driven medium-range weather forecasting, specifically 3–5 days.
Notably, our Weather2K is a spatio-temporal dataset with 20 optional multivariate meteorological factors, which can be applied to different directions of weather forecasting.


\begin{table*}[!th]\tiny
\caption{Definitions and physical descriptions of the variables in the Weather2K dataset.} \label{table1}
\begin{center}
\tabcolsep= 0.25cm
\begin{tabular}{l|c|c|l}
\midrule[1pt]
\textbf{Long Name} & \textbf{Short Name} &\textbf{Unit} & \textbf{Physical Description} \\ \midrule[1pt]
Latitude & lat & (°) & The latitude of the ground observation station \\ \hline
Longitude & lon & (°) & The longitude of the ground observation station \\ \hline
Altitude & alt & (m) & The altitude of the air pressure sensor \\ \midrule[1pt]
Air pressure & ap & hpa & Instantaneous atmospheric pressure at 2 meters above the ground \\ \hline
Water vapor pressure & wvp & hpa & Instantaneous partial pressure of water vapor in the air \\ \hline
Air Temperature & t & (°C) &  Instantaneous temperature of the air at 2.5 meters above the ground where sheltered from direct solar radiation \\ \hline
Maximum / Minimum temperature & mxt / mnt & (°C) & Maximum / Minimum air temperature in the last one hour \\ \hline
Dewpoint temperature & dt & (°C) & Instantaneous temperature at which the water vapor saturates and begins to condense\\ \hline
Land surface temperature & st & (°C) & Instantaneous temperature of bare soil at the ground surface \\ \hline
Relative humidity & rh & (\%) & Instantaneous humidity relative to saturation at 2.5 meters above the ground\\ \hline
Wind speed & ws & ($\mathrm{ms^{-1}}$) & The average speed of the wind at 10 meters above the ground in a 10-minute period  \\ \hline
Maximum wind speed & mws & ($\mathrm{ms^{-1}}$) & Maximum wind speed in the last one hour \\ \hline
Wind direction & wd & (°) & The direction of the wind speed. (Wind direction is 0 if wind speed is less than or equal to 0.2)\\ \hline
Maximum wind direction & mwd & (°) & Maximum wind speed's direction in the last one hour \\ \hline
Vertical visibility & vv & (m) & Instantaneous vertical visibility \\ \hline
Horizontal visibility in 1 min / 10 min & hv1 / hv2 & (m) & 1 / 10 minute(s) mean horizontal visibility at 2.8 meters above the ground \\ \hline
Precipitation in 1h / 3h / 6h / 12h / 24h  & p1 / p2 / p3 / p4 / p5 & (mm) & Cumulative precipitation in the last 1 / 3 / 6 / 12 / 24 hour(s)\\ \midrule[1pt]
\end{tabular}
\end{center}
\end{table*}


\subsection{Transformers in Time Series Forecasting}
Due to the special characteristics of time series forecasting task, the innovation of Transformer \citep{vaswani2017attention} and its variants based on the self-attention mechanism shows great capability in sequential data.
MetNet \citep{sonderby2020metnet} uses axial self-attention to aggregate the global context from radar and satellite data for precipitation forecasting.
LogTrans \citep{li2019enhancing} and Reformer \citep{kitaev2020reformer} propose the LogSparse attention and the local-sensitive hashing attention to reduce the complexity of both memory and time, respectively.
AST \citep{wu2020adversarial} utilizes a generative adversarial encoder-decoder pipeline.
Autoformer \citep{wu2021autoformer} proposes a seasonal-trend decomposition architecture with an auto-correlation mechanism.
Informer \citep{zhou2021informer} utilizes Kullback-Leibler divergence based the ProbSparse attention.
Pyraformer \citep{liu2021pyraformer} captures different ranges of temporal dependencies in a compact, multi-resolution fashion.
FEDformer \citep{zhou2022fedformer} uses Fourier transform and wavelet transform to consider the characteristics of time series in the frequency domain.

\subsection{Spatio-temporal Graph Neural Networks.}
Spatio-temporal forecasting models are mostly based on GNNs due to their ability to learn representations of spatial irregular distributed signals by aggregating or diffusing messages from or to neighborhoods.
GNNs have already brought a huge boost in traffic forecasting, such as STGCN \citep{yu2017spatio}, DCRNN \citep{li2017diffusion}, MSTGCN \citep{guo2019attention}, ASTGCN \citep{guo2019attention}, TGCN \citep{zhao2019t}, AGCRN \citep{bai2020adaptive}, and GMAN \citep{zheng2020gman}.
Moreover, LRGCN \citep{li2019predicting}, 2s-AGCN \citep{shi2019two} and MPNN \citep{panagopoulos2021transfer} highlight the usefulness of GNNs in path failure in a telecommunication network, skeleton-based action recognition and epidemiological prediction, respectively.
In the field of meteorological science, DeepSphere\citep{defferrard2020deepsphere} introduces a method based on a graph representation of the sampled spherical meteorological data.
PM2.5-GNN \citep{wang2020pm2} and CLCRN \citep{lin2022conditional} are used to predict PM2.5 concentrations and meteorological factors including temperature, cloud cover, humidity, and surface wind component.
In general, researches on the application of GNNs to weather forecasting are relatively scarce.

\section{THE DATASET OF WEATHER2K}

The raw data comes from CMA's observation data of China's ground weather stations.
Observation instruments and sensors of different weather stations all follow the standard of \emph{Specifications for surface meteorological observation—General} (GB/T 35221-2017) and \emph{Quality control of surface meteorological observation data} (QX/T 118-2010).
While collecting the original data, basic quality control (QC) technologies are first used for different meteorological factors to correct for sensor noise, bias, and external factors.
Focusing on missing values, defaults, outliers, we performed rigorous filtering and processing to ensure the quality of our dataset.
The main steps of our data processing procedure are provided in Supplementary Material C.1.

\subsection{Dataset Statistics}

\begin{figure}[t] 
\centerline{\includegraphics[scale=0.29]{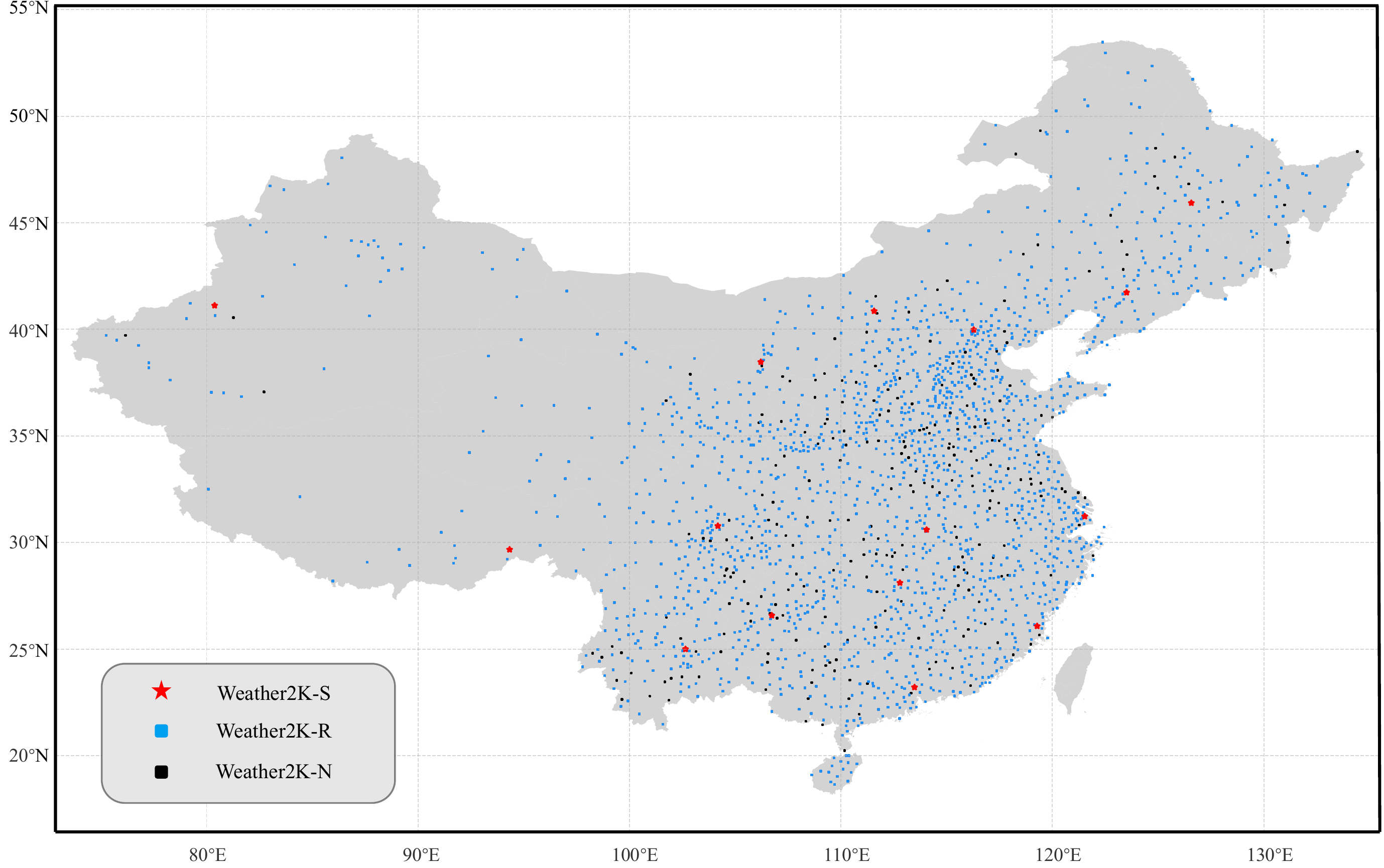}}
\caption{The distribution of the 2,130 ground weather stations that supply observation data for Weather2K.}
\label{f1}
\end{figure}

Weather2K contains the observation data from 2,130 ground weather stations throughout China, covering an area of more than 6 million square kilometers.
The data are available from January 2017 to August 2021 with a temporal resolution of 1 hour, in which we can ensure the use of unified meteorological observation instruments and sensors for data collection.
All stations use CST time (UTC + 8h).
Based on the consideration of building a representative and comprehensive dataset for meteorological forecasting, Weather2K contains 3 time-invariant constants: latitude, longitude, and altitude to provide position information and 20 important meteorological factors with a length of 40,896 time steps.
The definitions and physical descriptions of the chosen variables are listed in Table~\ref{table1}.
The current version of our open source Weather2K is Weather2K-R, which contains 1866 ground weather stations and is stored in Numpy format.
Weather2K-N, which contains the rest of the stations, is temporarily reserved for confidentiality reasons.
In addition, we further provide a special version of the Weather2K (Weather2K-S), which contains 15 representative ground weather stations distributed in different regions and is stored in CSV format files.
As shown in Figure~\ref{f1}, the red stars, blue squares, and black squares represent the stations that make up Weather2K-S, Weather2K-R, and Weather2K-N, respectively.
See more detailed statistics on Weather2K in Supplementary Material C.2.

\subsection{Comparison with Other Datasets}

Our Weather2K is a novel attempt to solve the task of weather forecasting by entirely using the observation data from the ground weather stations.
Although CMFD \citep{he2020first} also uses approximately 700 stations of CMA as its primary data source, it must incorporate several gridded datasets from remote sensing and reanalysis.
Jena Climate (https://www.bgc-jena.mpg.de/wetter/) only records the data of 1 weather station in Jena, Germany.

Most of the existing multivariate datasets cover a long time range. For example, CMFD \citep{he2020first} and WeatherBench \citep{rasp2020weatherbench} are both from 1979 to 2018. Limited by the purpose of data collection using unified meteorological observation instruments and sensors, our Weather2K begins in January 2017 and is ongoing.

The most attractive advantage of our Weather2K is the real-time and reliability of the data.
The combination of high precision sensor and high-quality data control ensures the accuracy of the data to the greatest extent.
The efficient and practical data processing can obtain hourly real-time data.
In comparison, CMFD \citep{he2020first}, ERA5 \citep{hersbach2020era5}, and WeatherBench \citep{rasp2020weatherbench} have a latency of at least 3 months because they are limited by the release time of the raw data.

In terms of meteorological factors, our Weather2K provides a total number of 20 optional variables.
By contrast, Climate Change (http://berkeleyearth.org/data/) focuses on meteorological factors related to temperature.
CMFD \citep{he2020first} provides 7 near-surface meteorological factors.
WeatherBench \citep{rasp2020weatherbench} contains 14 meteorological factors, 8 of which have multiple vertical levels.
Moreover, our Weather2K can be applied to weather forecasting tasks in different directions. Weather2K-S and Weather2K-R are designed for the task of time series forecasting and spatio-temporal forecasting, respectively, while existing datasets mainly focus on solving a single task.

\section{APPLICATIONS}

In this section, we mainly introduce the application of the Weather2K in two important directions of weather forecasting, namely time series forecasting and spatio-temporal forecasting.
For the convenience of expression, the temperature, visibility, and humidity mentioned in section 4 and section 5 refer to air temperature, horizontal visibility in 10 min, and relative humidity, respectively.
These 3 meteorological factors are also selected as our forecasting targets.
See more details about baseline models and implementation in Supplementary Material D.1 and D.2, respectively.

\begin{table*}[!ht]
\tiny
\caption{Univariate results and multivariate results under different baselines with different prediction lengths of 24, 72, 168, 336, and 720 steps. The input time length is 72 steps. The reported results of mean and standard deviation are obtained through experiments under 15 different ground weather stations in Weather2K-S. Results with \textbf{bold} and \underline{underlines} are the best and worst performance achieved by baselines, respectively.} \label{table2}

\begin{center}
\tabcolsep= 0.22cm
\renewcommand\arraystretch{0.8}
\begin{tabular}{c|c|c c|c c|c c|c c }
\midrule[1pt]
\multicolumn{2}{c|}{\textbf{Models}} & \multicolumn{2}{c|}{Transformer}& \multicolumn{2}{c|}{Reformer} & \multicolumn{2}{c|}{Informer} & \multicolumn{2}{c}{Autoformer} \\ \midrule[1pt]

\textbf{Factors} & \textbf{Metrics} & MSE & MAE & MSE & MAE & MSE & MAE & MSE & MAE  \\ \midrule[1pt]

\multirow{5}{*}{Temperature} & 24 & \textbf{0.0740}$\pm$0.0214 & \textbf{0.2000}$\pm$0.0284 & 0.0897$\pm$0.0241 & 0.2306$\pm$0.0289 & 0.0758$\pm$0.0229 & 0.2031$\pm$0.0299 & \underline{0.0978}$\pm$0.0290 & \underline{0.2367}$\pm$0.0342 \\ 
& 72 & \textbf{0.1246}$\pm$0.0396 & \textbf{0.2656}$\pm$0.0440 & 0.1294$\pm$0.0393 & 0.2797$\pm$0.0400 & 0.1272$\pm$0.0355 & 0.2711$\pm$0.0386 & \underline{0.1384}$\pm$0.0474 & \underline{0.2832}$\pm$0.0464  \\ 
& 168 & 0.1560$\pm$0.0513 & 0.2999$\pm$0.0502 & \textbf{0.1431}$\pm$0.0528 & \textbf{0.2926}$\pm$0.0529 & \underline{0.1699}$\pm$0.0529 & \underline{0.3162}$\pm$0.0660 & 0.1574$\pm$0.0583 & 0.3002$\pm$0.0544 \\ 
& 336 & 0.1709$\pm$0.0635 & 0.3148$\pm$0.0587 & \textbf{0.1451}$\pm$0.0562 & \textbf{0.2947}$\pm$0.0576 & \underline{0.2118}$\pm$0.0660 & \underline{0.3586}$\pm$0.0508 & 0.1655$\pm$0.0630 & 0.3092$\pm$0.0575 \\
& 720& 0.1603$\pm$0.0671 & 0.3058$\pm$0.0630 & \textbf{0.1481}$\pm$0.0532 & \textbf{0.2968}$\pm$0.0544 & \underline{0.2119}$\pm$0.0665 & \underline{0.3637}$\pm$0.0536 & 0.2011$\pm$0.0602 & 0.3453$\pm$0.0510 \\\midrule[1pt]

\multirow{5}{*}{Visibility} & 24 & 0.7105$\pm$0.1155 & 0.6536$\pm$0.1151 & \textbf{0.6623}$\pm$0.1108 &\textbf{ 0.6288}$\pm$0.1175 & 0.7007$\pm$0.1198 & 0.6390$\pm$0.1218 & \underline{0.8014}$\pm$0.1589 & \underline{0.6929}$\pm$0.1013 \\ 
& 72 & 0.8722$\pm$0.1945 & 0.7475$\pm$0.1487 & \textbf{0.8304}$\pm$0.1843 & \textbf{0.7440}$\pm$0.1526 & 0.8847$\pm$0.2079 & 0.7503$\pm$0.1400 & \underline{0.9350}$\pm$0.2215 & \underline{0.7585}$\pm$0.1336  \\ 
& 168 & 0.9153$\pm$0.1969 & 0.7802$\pm$0.1439 & \textbf{0.9083}$\pm$0.2380 & \textbf{0.7795}$\pm$0.1441 & 0.9428$\pm$0.1942 & \underline{0.7943}$\pm$0.1523 & \underline{0.9911}$\pm$0.2273 & 0.7834$\pm$0.1289  \\ 
& 336 & 0.9952$\pm$0.2766 & 0.8059$\pm$0.1855 & \textbf{0.9219}$\pm$0.2713 & \textbf{0.7810}$\pm$0.1706 & \underline{1.0232}$\pm$0.2336 & \underline{0.8388}$\pm$0.1720 & 0.9919$\pm$0.2311 & 0.7832$\pm$0.1415  \\
& 720 & 0.9963$\pm$0.2570 & 0.8090$\pm$0.1755 & \textbf{0.9506}$\pm$0.2853 & \textbf{0.7946}$\pm$0.1763 & 1.0235$\pm$0.2618 & \underline{0.8438}$\pm$0.0789 & \underline{1.0651}$\pm$0.2591 & 0.8145$\pm$0.1553 \\ \midrule[1pt]

\multirow{5}{*}{Humidity} & 24 & 0.3746$\pm$0.0820 & 0.4607$\pm$0.0548 & \textbf{0.3618}$\pm$0.0725 & \textbf{0.4529}$\pm$0.0438 & 0.3749$\pm$0.0848 & 0.4563$\pm$0.0529 & \underline{0.4291}$\pm$0.1058 & \underline{0.4941}$\pm$0.0625 \\ 
& 72 & 0.5228$\pm$0.1310 & 0.5603$\pm$0.0782 & \textbf{0.4932}$\pm$0.1074 & \textbf{0.5455}$\pm$0.0621 & 0.5232$\pm$0.1234 & 0.5619$\pm$0.0769 & \underline{0.5511}$\pm$0.1209 & \underline{0.5721}$\pm$0.0651  \\ 
& 168 & 0.5935$\pm$0.1533 & 0.6004$\pm$0.0814 & \textbf{0.5421}$\pm$0.1254 & \textbf{0.5759}$\pm$0.0688 & \underline{0.6221}$\pm$0.1567 & \underline{0.6163}$\pm$0.0884 & 0.6111$\pm$0.1304 & 0.6042$\pm$0.0651 \\ 
& 336 & 0.6107$\pm$0.1643 & 0.6142$\pm$0.0815 & \textbf{0.5761}$\pm$0.1456 & \textbf{0.5938}$\pm$0.0749 & \underline{0.6965}$\pm$0.1845 & \underline{0.6653}$\pm$0.1051 & 0.6514$\pm$0.1567 & 0.6258$\pm$0.0737  \\
& 720& 0.6177$\pm$0.1638 & 0.6146$\pm$0.0833 & \textbf{0.5864}$\pm$0.1399 & \textbf{0.6009}$\pm$0.0737 & \underline{0.7979}$\pm$0.1538 & \underline{0.7285}$\pm$0.0789 & 0.7145$\pm$0.1631 & 0.6599$\pm$0.0741 \\\midrule[1pt]\midrule[1pt]

\multirow{5}{*}{Multivariate} & 24 & 0.7134$\pm$0.1796 & 0.4597$\pm$0.0418 & \textbf{0.6342}$\pm$0.1700 & \textbf{0.4177}$\pm$0.0178 & 0.6838$\pm$0.1777 & 0.4381$\pm$0.0338 & \underline{0.7525}$\pm$0.1968 & \underline{0.4868}$\pm$0.0300 \\ 
& 72 & 0.8143$\pm$0.2071 & 0.5179$\pm$0.0469 & \textbf{0.7616}$\pm$0.1877 & \textbf{0.4914}$\pm$0.0330 & 0.8142$\pm$0.1967 & 0.5128$\pm$0.0391 & \underline{0.8585}$\pm$0.2132 & \underline{0.5390}$\pm$0.0314  \\ 
& 168 & 0.8462$\pm$0.2076 & 0.5261$\pm$0.0430 & \textbf{0.8029}$\pm$0.1979 & \textbf{0.5075}$\pm$0.0339 & 0.8641$\pm$0.2004 & \underline{0.5543}$\pm$0.0311 & \underline{0.8945}$\pm$0.2160 & 0.5455$\pm$0.0390 \\ 
& 336 & 0.8507$\pm$0.2044 & 0.5450$\pm$0.0396 & \textbf{0.8251}$\pm$0.2057 & \textbf{0.5173}$\pm$0.0337 & 0.9040$\pm$0.2129 & \underline{0.5750}$\pm$0.0307 & \underline{0.9298}$\pm$0.2222 & 0.5575$\pm$0.0381  \\
& 720& 0.8967$\pm$0.2308 & 0.5543$\pm$0.0446 & \textbf{0.8501}$\pm$0.2279 & \textbf{0.5240}$\pm$0.0335 & 0.9242$\pm$0.2241 & \underline{0.5833}$\pm$0.0329 & \underline{0.9379}$\pm$0.2553 & 0.5731$\pm$0.0432 \\\midrule[1pt]

\end{tabular}
\end{center}
\end{table*}

\begin{table*}[!ht]\scriptsize
\caption{Univariate results with 8 spatio-temporal GNN models in prediction time length of 12 steps on Weather2K-R. The input time length is 12 steps. Results with \textbf{bold} and \underline{underlines} are the best and worst performance, respectively.} 
\label{table3}
\begin{center}
\tabcolsep= 0.4cm
\renewcommand\arraystretch{0.8}
\begin{tabular}{c|c|c c c c c c c c}

\midrule[1pt]

\textbf{Factors} & \textbf{Metrics} & STGCN & ASTGCN &  MSTGCN & DCRNN& GCGRU& TGCN &  A3TGCN&  CLCRN\\ \midrule[1pt]

\multirow{2}{*}{Temperature} & MAE & 3.1297 & 2.6942 & 3.6827 & \underline{5.0457} & 1.8823 & 2.6429 & 2.9372 & \textbf{1.6498} \\ 
& RMSE & 4.7154 & 3.6383 & 4.9195 & \underline{6.4070} & 2.7491 & 3.6189 & 4.0045 & \textbf{2.4328}  \\ 
\midrule[1pt]

\multirow{2}{*}{Visibility} & MAE & \underline{7.7568} & 5.7311 & 5.9053 & 7.6464 & 4.4863 & 6.0115 & 5.7731 & \textbf{4.1252} \\ 
& RMSE & \underline{9.0896} & 7.5523 & 7.8016 & 9.0035 & 6.8236 & 8.4893 & 7.5974 & \textbf{6.4660} \\ 
\midrule[1pt]

\multirow{2}{*}{Humidity} & MAE & 9.8479 & 12.0332 & 14.4780 & \underline{16.3467} & \textbf{7.7008} & 10.7045 & 12.1034 & 7.8056 \\ 
& RMSE & 14.5123 & 15.9847 & 18.7188 & \underline{21.2517} & \textbf{10.9422} & 14.3170 & 15.9838 & 11.1326  \\ 
\midrule[1pt]

\end{tabular}
\end{center}
\end{table*}

\subsection{Time Series Forecasting}
Weather2K-S is the special version of Weather2K designed for the task of time series forecasting, which contains 15 representative and geographically diverse ground weather stations distributed throughout China.
We conduct three types of experiments on Weather2K-S: univariate forecasting, multivariate forecasting, and multivariate to univariate forecasting.
4 representative transformer-based baselines are set up for comparison, i.e., Transformer \citep{vaswani2017attention}, Reformer \citep{kitaev2020reformer}, Informer \citep{zhou2021informer}, and Autoformer \citep{wu2021autoformer}.
In meteorological science, short-term, medium-term, and long-term forecasting can be divided into 0-3 days, 3-15 days, and more than 15 days, respectively.
For better comparison, we set the prediction time lengths for both training and evaluation to 24 (1 day), 72 (3 days), 168 (7 days), 336  (14 days), and 720 (30 days) steps, while the input time length is fixed to 72 (3 days) steps.
We evaluate the performance of predictive models by two widely used metrics, including Mean Square Error (MSE) and Mean Absolute Error (MAE).

Table~\ref{table2} shows the forecasting performance of temperature, visibility, and humidity under different baselines and prediction lengths.
In univariate forecasting and multivariate forecasting, both the input and output are the single and multiple meteorological factors we specify, respectively.
By comparison, we can intuitively find that multivariate forecasting is more complex than univariate forecasting.
Meteorological forecasting becomes more challenging as the length of the predicted time series increases.
On Weather2K-S, Reformer \citep{kitaev2020reformer} have the best performance most of the time due to its efficient locality-sensitive hashing attention.
The poor performance of Autoformer \citep{wu2021autoformer} may be that it attempts to construct a series-level connection based on the process similarity derived by series periodicity, whereas meteorological data in Weather2K-S change dynamically and irregularly.

We further explore to use the multivariate data to forecast the target univariate meteorological factor. 
However, multivariate information is underutilized in existing transformer-based time series forecasting baselines.
The performance of multivariate to univariate forecasting is provided in Supplementary Material D.3.
To make the evaluation more complete and informative, we also present the univariate forecasting results of the persistent model and some classical nonparametric methods in Supplementary Material D.4.

\begin{figure*}[!ht] 
\centerline{\includegraphics[scale=0.62]{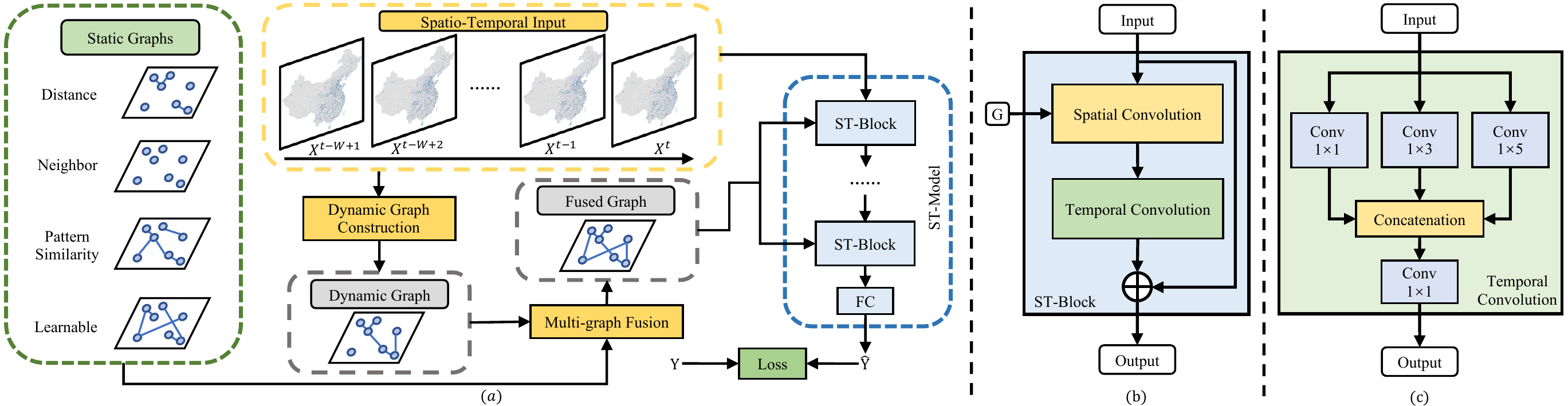}}
\caption{(a) The overview of Meteorological Factors based Multi-Graph Convolution Network (MFMGCN). (b) The architecture of spatio-temporal block (ST-block). (c) The architecture of the multi-branch temporal convolution.}
\label{f2}
\end{figure*}


\subsection{Spatio-Temporal Forecasting}

Unlike time series forecasting that only targets a single selected ground weather station, spatio-temporal forecasting that can cover the entire region is more challenging and meaningful in the meteorological domain due to the highly nonlinear temporal dynamics and complex location-characterized patterns.
We conduct experiments for univariate forecasting on Weather2K-R.
We select 8 state-of-the-art spatio-temporal GNN models as baselines: STGCN \citep{yu2017spatio}, ASTGCN \citep{guo2019attention}, MSTGCN \citep{guo2019attention}, DCRNN \citep{li2017diffusion}, GCGRU \citep{seo2018structured}, TGCN \citep{zhao2019t}, A3TGCN \citep{bai2021a3t}, and CLCRN \citep{lin2022conditional}.
We construct a neighbor graph based on distance as one of the GNN models' inputs by utilizing three time-invariant constants that provide position information including latitude, longitude, and altitude.
We set both the input time length and prediction time length as 12 steps.
We evaluate the performance by two commonly used metrics in spatio-temporal forecasting, including MAE, Root Mean Square Error (RMSE).

Table~\ref{table3} summarizes the univariate forecasting results of 8 spatio-temporal GNN models based on Weather2K-R.
Since most of the compared models are established for the task of traffic forecasting, it is not surprising that they show significant degradation in performance for meteorological forecasting, such as DCRNN \citep{li2017diffusion} and MSTGCN \citep{guo2019attention}.
Moreover, the performance of 8 spatio-temporal GNN models in forecasting different meteorological factors varies widely.
For example, STGCN \citep{yu2017spatio} ranks third in humidity forecasting, but worst in visibility forecasting.
CLCRN \citep{lin2022conditional} achieves the best performance because the conditional local kernel is specifically designed to capture the meteorological local spatial patterns.
In addition, the current GNN baselines focus on the univariate forecasting and do not consider correlated multiple variables as cooperative inputs, which may have important impacts on weather forecasting.

Therefore, there are two urgent problems in the application of spatio-temporal forecasting in the meteorological domain: (1) A novel and general GNN framework designed specifically for the task of weather forecasting. (2) Make effective use of the multivariate meteorological factors.

\section{MFMGCN}
It is insufficient to simply use distance similarity to represent the correlations among observed data with spatio-temporal attributes \citep{geng2019spatiotemporal}.
Modeling with a single graph usually brings unexpected bias, while multiple graphs can attenuate and offset the bias.
To solve the above problems, we introduce MFMGCN, which combines several static graphs with the dynamic graph to effectively construct the intrinsic correlation among geographic locations based on meteorological factors.
The framework of MFMGCN is shown in Figure~\ref{f2}.
Through extensive experiments and ablation studies, we have demonstrated that MFMGCN achieves considerable forecasting performance gain and shows great robustness in the temporal dimension compared with previous methods on Weather2K-R.

\subsection{Problem Setting} 
In this section, we describe the problem setting of the spatio-temporal weather forecasting.
The meteorological network can be represented by a graph $G =\left(V\text{, } E\text{, } \mathbf{A}  \right) $,
where $V$ denotes the set of locations, $E$ is a set of edges indicating the connectivity between the nodes and $\mathbf A \in  \mathbf{R}^{N \times N} $is the adjacency matrix.
Let $\mathbf{X}^{\left(t\right)} \in \mathbf{R}^{N \times D} $ represent the value of meteorological factors in all locations at the $t$-th interval,where $N$ and $D$ is the number of locations and factors, respectively.
The spatio-temporal weather forecasting problem is to learn a function $P$ that maps historical $W^{'}$ observed signals to the future $W$ signals:
\begin{equation}
\resizebox{0.9\hsize}{!}{$
\left[\mathbf{X}^{\left(t-W^{\prime}+1\right)}, \ldots, \mathbf{X}^{(t)} ;{G}\right] \stackrel{P}{\longrightarrow}\left[\mathbf{X}^{(t+1)}, \ldots, \mathbf{X}^{(t+W)} ; {G}\right]
$}
\end{equation}
We set $D$ = 1 for univariate forecasting in this study.
We set $W^{'}$ = $W$ = 12, which means
both the input time length and forecasting time length are 12 steps.

\subsection{Static Graph Construction} 
In this section, we describe in detail 4 static graphs used in MFMGCN that represent different types of correlations among geographic locations, including:
(1) The distance graph $ G_D=\left(V\text{, } E\text{, } \mathbf{A}_D \right) $.
(2) The neighbor graph $ G_N=\left(V\text{, } E\text{, } \mathbf{A}_N \right) $.
(3) The pattern similarity graph $ G_P=\left(V\text{, } E\text{, } \mathbf{A}_P \right) $.
(4) The learnable graph $ G_L=\left(V\text{, } E\text{, } \mathbf{A}_L \right) $.

\subsubsection{Distance Graph}
The distance-based graph describes the topological structure of the ground weather station network.
Our distance graph $G_D$ is constructed based on the spherical distance of stations' spatial location with thresholded Gaussian kernel \citep{shuman2013emerging}. The element of distance matrix $\mathbf{A}_D$ is defined as follows:

\begin{equation}
\resizebox{0.9\hsize}{!}{$
A_{D, i j}= \begin{cases}\exp \left(-\frac{d_{i j}^2}{\sigma_D^2}\right) & \text {, for } i \neq j \text { and } \exp \left(-\frac{d_{i j}^2}{\sigma_D^2}\right) \geq \varepsilon \\ 0 & \text {, otherwise}\end{cases}
$}
\end{equation}

where \(d_{i j}\) represents the spherical distance between \(v_{i}\) and \(v_{j}\). $\varepsilon$ and $\sigma_D^2$ are used to control the sparsity and distribution of $\mathbf{A}_D$, respectively.

\subsubsection{Neighbor Graph}
We construct the neighbor graph $G_N$ by simply connecting a location to its adjacent $N_A$ stations.
The element of the neighbor matrix $\mathbf{A}_N$ is defined as follows:
\begin{equation}
A_{N, i j}= \begin{cases}1 &\text {, } v_i \text { and } v_j \text { are adjacent } \\ 0 &\text {, }  \text {otherwise}\end{cases}
\end{equation}
The influence of the selection value of $N_A$ is discussed in Supplementary Material D.5.

\subsubsection{Pattern Similarity Graph}
Since distant stations may also have highly consistent meteorological characteristics, it is essential to establish a graph by mining pattern similarity between nodes based on our multivariate dataset.
The element of pattern similarity matrix $\mathbf{A}_P$ is defined as follows:
\begin{equation}
\resizebox{0.9\hsize}{!}{$
A_{P, i j}^f= \begin{cases}\frac{\sum_{p=1}^P\left(t_{i, p}^f-\overline{T_i^f}\right)\left(t_{j, p}^f-\overline{T_j^f}\right)}{\sqrt{\sum_{i=1}^p\left(t_{i, p}^f-\overline{T_i^f}\right)^2} \sqrt{\sum_{j=1}^p\left(t_{j, p}^f-\overline{T_j^f}\right)^2}} & \text {, if } i \neq j \\ 0 & \text {, otherwise }\end{cases}
$}
\end{equation}

Note that the pattern similarity graph can be constructed using Pearson correlation coefficients \citep{zhang2020crowd} based on any meteorological factor $f \in F$, where $F$ denotes the set of all factors in Weather2K.
For a specific feature 
$f$, $T_i^f=\left\{t_{i, 1}^f, t_{i, 2}^f, \cdots, t_{i, p}^f, \cdots, t_{i, P}^f\right\}$ 
describe the time-series sequence of $v_{i}$ used for training, 
where $P$ is the length of the series, and $t_{i, p}^f$ is the time-series data value of the vertex $v_{i}$ at time step $p$.
To model the correlation between any geographic location comprehensively, we build pattern graphs on temperature, visibility, and humidity.

\subsubsection{Learnable Graph}
The meteorological data need not only the accumulation of long time series for historical vertical comparative study, but also the horizontal comparative study of global regions.
Despite the above graphs can capture spatial dependencies in detail, the information of the node itself is not sufficiently utilized due to the lack of historical level and geographic information which is challenging to predefine.
Inspired by \citet{wu2020connecting}, we propose static node embedding 
$\mathbf{E} \in \mathbf{R} ^ D$
which is learnable to encode meteorological statistics for each node and compute:
\begin{equation}
\mathbf{M}_i=\tanh \left(\alpha \mathbf{E}_i \mathbf\Theta_1\right)
\end{equation}
\begin{equation}
\mathbf{M}_j=\tanh \left(\alpha \mathbf{E}_j \mathbf\Theta_2\right)
\end{equation}
\begin{equation}
{A}_{L, i j}=\operatorname{ReLU}\left(\tanh \left(\alpha\left(\mathbf{M}_i\cdot \mathbf{M}_j-\mathbf{M}_j \cdot\mathbf{M}_i\right)\right)\right)
\end{equation}
where $\mathbf{E}_i$ and $\mathbf{E}_j$ represents randomly initialized node embeddings of node $v_i$ and $v_j$, respectively. $ \mathbf\Theta_1$,$ \mathbf\Theta_2$ are linear layers and $ \alpha$ is a hyper-parameter for controlling the saturation rate of the activation function.
To make $G_L$ more applicable to the meteorological field, each node of $G_L$ has more connections with other nodes in different geographic locations, rather than selecting only the top-k closest nodes as its connections.
And the edges of $G_L$ are not limited to uni-directional because the geographical situation has two-way effects under different conditions.

\subsection{Dynamic Graph Construction} 
In this section, the construction of the dynamic graph is presented. 
A node may have different types and degrees of weather, leading to dynamic changes in the relationship with other nodes. However, it is still hard to model such dynamic characteristics by static adjacency matrix based on distance or neighbor.
Attention mechanism methods make computation and memory grow quadratically with the increase of graph size while RNN-based methods are limited by the length of the input sequence. 
Both of them need to be carefully modified to ensure successful training on Weather2K-R.
We dynamically construct a graph $G_K=\left(V,E,\mathbf{A}_K \right) $ to model nonlinear spatio-temporal correlations in a efficient way.

Note that $ \mathbf{X}_{t,i}=\left\{x_{t-W+1,i},\cdots,x_{t-1,i},\cdots, x_{t,i}\right\}\in \mathbf{R} ^ {W \times D}$ 
is an input temporal sequence of vertex $v_i$ at time $t$, where $W$ is the sequence length and D is the number of factors.
We then simply flatten the vector to $\mathbf{Z}_i \in \mathbf{R} ^ {WD}$ to represent the characteristics on the input temporal scale. 
Similar to the way of building the learnable graph, the element of matrix $\mathbf{A}_K$ is defined as follows:
\begin{equation}
\mathbf{D}_i=\tanh \left(\beta \mathbf{Z}_i \mathbf{W}_1\right)
\end{equation}
\begin{equation}
\mathbf{D}_j=\tanh \left(\beta \mathbf{Z}_j \mathbf{W}_2\right)
\end{equation}
\begin{equation}
{A}_{K, i j}=\operatorname{ReLU}\left(\tanh \left(\beta\left(\mathbf{D}_i\cdot \mathbf{D}_j-\mathbf{D}_j \cdot\mathbf{D}_i\right)\right)\right)
\end{equation}
where $\mathbf W_1$ and $\mathbf W_2$ are linear layers and $\mathbf \beta$ is a hyper-parameter for controlling the saturation rate of the activation function.
Different from the learnable graph based on static node embedding, $G_K$ is generated dynamically both in the training and inference stages.

\subsection{Multi-graph Fusion} 
Since different graphs contain specific spatio-temporal information and make unequal contributions to the forecasting result of every single node, it is inappropriate to simply merge them using weighted sum or other averaging approaches in a unified way. 
Therefore, we use learnable weights $\mathbf{W}_s \in \mathbf{R}^{N \times N} $ to describe the importance of graph $s$ for each node, where $N$ presents the number of nodes and $s \in S=\{D,N,P,L,K\}$ is one of the proposed graphs.
Finally, the result of multi-graph fusion is $G_{fused}=\left(V,E,\mathbf{A}_{fused}\right)$:
\begin{equation}
{\mathbf{A}_{fused}}=\sum_{s \in S}\mathbf{W}_s \odot {\mathbf{A}}_s
\end{equation}
where $\odot$ indicates the element-wise Hadamard product.

\subsection{Spatio-Temporal Graph Neural Network} 
Considering the time-space consistency is important in weather forecasting, we build a model by stacking spatio-temporal blocks (ST-block) which are complete convolutional structures.
As shown in Figure~\ref{f2} (b),
the ST-block adopts a residual learning framework which consists of a spatial convolution and a temporal convolution in order.
Then, a fully-connected output layer is used to generate the final prediction $\hat{\mathbf{Y}}$.

\subsubsection{Graph Convolution in Spatial Dimension}
Following \citet{yu2017spatio}, we adopt the spectral graph convolution to directly process the signals at each time slice. 
In spectral graph analysis, taking the weather conditions of factor $f$ at time $t$ as an example, the signal all over the graph is $x = \mathbf{x}_t^f \in \mathbf{R}^N $, then the signal on the graph $G$ is filtered by a kernel $g_\theta $:
\begin{equation}
\resizebox{0.85\hsize}{!}{$
g_\theta *_G x=g_\theta(\mathbf{L}) x=g_\theta\left(\mathbf{U} {\mathbf\Lambda} \mathbf{U}^T\right) x=\mathbf{U} g_\theta({\mathbf\Lambda}) \mathbf{U}^T x
$}
\end{equation}
where graph Fourier basis 
$\mathbf{U} \in \mathbf{R}^{N \times N} $ 
is the matrix of eigenvectors of the normalized graph Laplacian 
$ \mathbf L=\mathbf I_N - \mathbf D^{-\frac{1}{2}} \mathbf A \mathbf D^{-\frac{1}{2}}=\mathbf U \mathbf{\Lambda} \mathbf U^T \in \mathbf{R}^{N \times N} $.
Note that $\mathbf I_N$ is an identity matrix and $\mathbf D \in \mathbf R^{N \times N} $ is the diagonal degree matrix with $\mathbf D_{ii}=\Sigma_j A_{i j}$ ($\mathbf A$ is the adjacent matrix). $\mathbf\Lambda \in \mathbf{R}^{N \times N}$ is the diagonal matrix of eigenvalues of $\mathbf L$ and filter $g_\theta(\mathbf{\Lambda})$ is also a diagonal matrix.
The above formula can be understood as Fourier transforming $g_\theta $ and $x$ respectively into the spectral domain, and multiplying their transformed results, and doing the inverse Fourier transform to get the output of the graph convolution \citep{shuman2013emerging}.

In practice, Chebyshev polynomials are adopted to solve the efficiency problem of performing the eigenvalue decomposition since the scale of our graph is large \citep{hammond2011wavelets}:
\begin{equation}
g_\theta *_G x=g_\theta(\mathbf{L}) x=\sum_{k=0}^{K-1} \theta_k T_k(\tilde{\mathbf{L}}) x
\end{equation}
where $T_k(\tilde{\mathbf{L}}) \in \mathbf{R}^{N \times N}$ is the Chebyshev polynomial of order $k$ evaluated at the scaled Laplacian $\tilde{\mathbf{L}}=2 \mathbf L / \lambda_{\max }-\mathbf I_N$, and $\lambda_{\max }$ denotes the largest eigenvalue of $\mathbf L$.
Note that $K$ is the kernel size of graph convolution, which determines the maximum radius of the convolution from central nodes.

\subsubsection{Convolution in Temporal Dimension}
Although RNN-based methods have superiority over CNN-based methods in capturing temporal dependency, the time-consuming problem of recurrent networks is nonnegligible, especially in the field of long-term weather forecasting.

In MFMGCN, a multi-branch convolution layer on the time axis is further stacked to model the temporal dynamic behaviors of meteorological data.
As shown in Figure~\ref{f2} (c),
the convolution layer of each branch has different receptive fields for extracting information at different scales.
Information of different scales is then fused through concatenation and a 1x1 convolution layer.

\subsubsection{Time-space Consistency}
The time-space consistency in weather forecasting means that the larger the time scale of the data, the wider the spatial range that can be predicted.
Inspired by this rule, we balance the span of space and time by setting appropriate parameters in ST-blocks.
The kernel size $K$ of graph convolution and receptive field in temporal convolution are increased linearly with the stacking of blocks.
By doing so, our model can integrate spatio-temporal information orderly and avoid imbalance between both sides.

\begin{table}[!b]\scriptsize
\tiny
\caption{The results of ablation studies.} \label{table4}

\begin{center}
\tabcolsep= 0.115cm
\renewcommand\arraystretch{1.8}
\begin{tabular}{c c c c c|c c|c c|c c}
\midrule[1pt]
\multirow{2}{*}{$G_D$} & \multirow{2}{*}{$G_N$} & \multirow{2}{*}{$G_P$} & \multirow{2}{*}{$G_L$} & \multirow{2}{*}{$G_K$} & \multicolumn{2}{c|}{Temperature}& \multicolumn{2}{c|}{Visibility}& \multicolumn{2}{c}{Humidity}\\
\cline{6-11}
& & & & & MAE & RMSE & MAE & RMSE & MAE & RMSE\\
\midrule[1pt]
\CheckmarkBold & & & & & 1.7190 & 2.5056 & 4.2625 & 6.3809 & 8.1178 & 11.6013 \\
 & \CheckmarkBold & & & & 1.7420 & 2.5211 & 4.2724 & 6.4342 & 8.0778 & 11.6008 \\
 & & \CheckmarkBold & & & 2.3651 & 3.1726 & 4.8104 & 6.5291 & 8.6860 & 12.2008 \\
 & & & \CheckmarkBold & & 1.5842 & 2.2753 & 4.1551 & 6.1609 & 7.2361 & 10.4721 \\
 & & & & \CheckmarkBold & 1.8267 & 2.5691 & 4.4877 & 6.3925 & 7.9877 & 11.5218 \\
 \CheckmarkBold & \CheckmarkBold & & & & 1.7430 & 2.5342 & 4.2403 & 6.3334 & 7.9565 & 11.4717 \\
 \CheckmarkBold & \CheckmarkBold & \CheckmarkBold & & & 1.6559 & 2.3834 & 4.1486 & 6.1512 & 7.7309 & 11.1712 \\
 & \CheckmarkBold & \CheckmarkBold & \CheckmarkBold & \CheckmarkBold & 1.5540 & 2.2448 & 3.9409 & 6.0163 & 7.1181 & 10.3577 \\
 \CheckmarkBold & & \CheckmarkBold & \CheckmarkBold & \CheckmarkBold & 1.5266 & 2.1925 & 3.9374 & 6.0278 & 7.1326 & 10.4074 \\
 \CheckmarkBold & \CheckmarkBold & & \CheckmarkBold & \CheckmarkBold & 1.5046 & 2.1769 & 3.9361 & 6.0685 & 7.2506 & 10.5775 \\
 \CheckmarkBold & \CheckmarkBold & \CheckmarkBold & & \CheckmarkBold & 1.5893 & 2.2887 & 4.4932 & 6.4680 & 7.4027 & 10.7293 \\
 \CheckmarkBold & \CheckmarkBold & \CheckmarkBold & \CheckmarkBold & & 1.5053 & 2.1677 & 4.0174 & 6.0701 & 7.1858 & 10.4025 \\
 \CheckmarkBold & \CheckmarkBold & \CheckmarkBold & \CheckmarkBold & \CheckmarkBold & \textbf{1.4418} & \textbf{2.0574} & \textbf{3.9277} & \textbf{6.0132} & \textbf{7.1125} & \textbf{10.3512} \\
\midrule[1pt]

\end{tabular}
\end{center}
\end{table}

\subsection{Experiments} 
\subsubsection{Ablation Study} 
In this section, we conduct ablation studies to show the effects of different fusion selections and discuss whether each graph works as intended in forecasting temperature, visibility, and humidity.
We first evaluate the performance of each graph separately.
Secondly, we set distance graph $G_D$ as the base input, which is the same as baseline models, to verify whether the stack of more graphs (i.e. more information) leads to better performance.
And then we compare the performance of all graphs but one (i.e. 4-graph fusion).
Finally, we show the performance of 5-graph fusion.

From Table~\ref{table4}, we can conclude that: (1) $G_D$ and $G_N$ is similar in results because they are both constructed based on location relationships, and it's better to combine the two graphs to make more efficient use of geographical distribution.
(2) Using $G_P$ separately yields a quite poor result.
However, $G_P$ brings a positive influence in multi-graph fusion.
(3) $G_L$ is the optimal single graph, and it also brings considerable performance gain in multi-graph fusion.
(4) Using $G_K$ separately does not work as well because it lacks the stable correlation provided by the static graph.
But it can be complementary to the static graph and brings expected gain.
(5) Whether forecasting temperature, visibility, or humidity, the effects of different fusion selections have similar trend on performance, which indicates that the multi-graph method is universal to meteorological factors.

\subsubsection{Overall Method Comparison} 

\begin{figure}[t] 
\centerline{\includegraphics[scale=0.42]{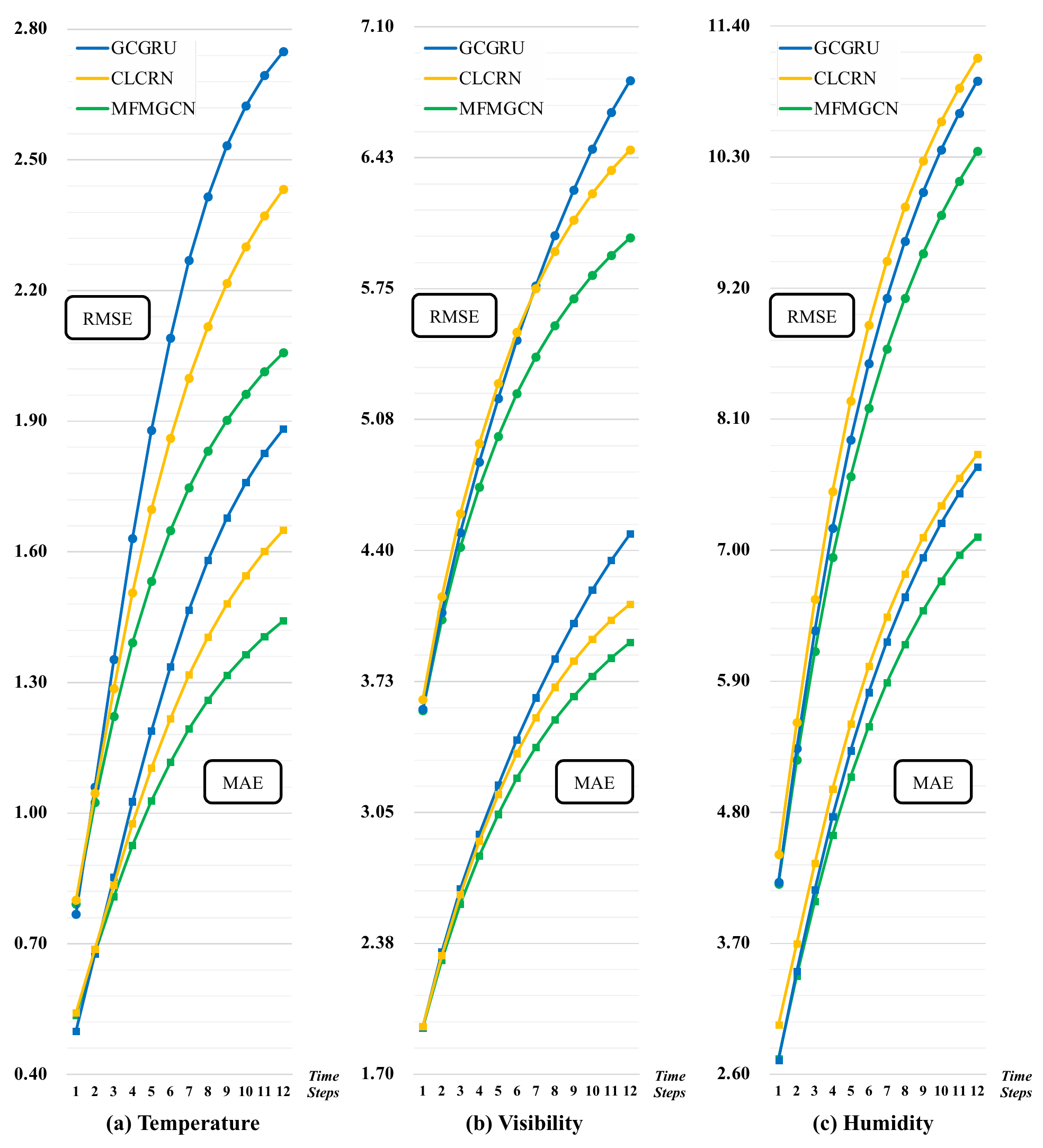}}
\vspace{.1in}
\caption{Overall method comparison in forecasting (a) temperature, (b) visibility, and (c) humidity.}
\label{f3}
\end{figure}

To avoid complicated and verbose plots, we give the comparison of methods with the overall top three performances on MAE and RMSE: GCGRU \citep{seo2018structured}, CLCRN \citep{lin2022conditional}, and our MFMGCN in Figure~\ref{f3}.
With the increase of the forecasting horizon from the first time step to 12 time steps, the gap between MFMGCN and other methods gradually widened.
It can be concluded that MFMGCN improves both the forecasting performance and temporal robustness, which is of great significance for real-world practical applications.
In addition, MFMGCN achieves the state-of-the-art performance in different meteorological factors, which further verifies it is a general framework for the task of weather forecasting.

\section{CONCLUSION}
This paper contributes a new benchmark dataset named Weather2K, which aims to make up for the deficiencies of existing weather forecasting datasets in terms of real-time, reliability, and diversity.
The data of Weather2K is hourly collected from 2,130 ground weather stations, which contain 20 meteorological factors and 3 constants for position information.
Weather2K is applicable to diverse tasks such as time series forecasting and spatio-temporal forecasting.
As far as we know, Weather2K is the first attempt to tackle weather forecasting task by taking full advantage of the strengths of observation data from ground weather stations.
Besides, we further propose MFMGCN, a novel and general framework designed specifically for the weather forecasting task, to effectively construct the intrinsic correlation among geographic locations based on meteorological factors by combining the static graph and the dynamic graph, which significantly outperforms previous spatio-temporal GNN methods on Weather2K.
We hope that the release of Weather2K can provide a foundation for accelerated research in this area and foster collaboration between atmospheric and data scientists.

\subsubsection*{Acknowledgements}
This work is partially supported by the MoE-CMCC “Artifical Intelligence” Project (Grant No. MCM20190701) and the BUPT innovation and entrepreneurship support program (Grant No. 2023-YC-A177).

\bibliographystyle{plainnat}
\bibliography{ms}

\begin{thebibliography}{48}
\providecommand{\natexlab}[1]{#1}
\providecommand{\url}[1]{\texttt{#1}}
\expandafter\ifx\csname urlstyle\endcsname\relax
  \providecommand{\doi}[1]{doi: #1}\else
  \providecommand{\doi}{doi: \begingroup \urlstyle{rm}\Url}\fi

\bibitem[Bai et~al.(2021)Bai, Zhu, Song, Zhao, Hou, Du, and Li]{bai2021a3t}
Jiandong Bai, Jiawei Zhu, Yujiao Song, Ling Zhao, Zhixiang Hou, Ronghua Du, and
  Haifeng Li.
\newblock A3t-gcn: Attention temporal graph convolutional network for traffic
  forecasting.
\newblock \emph{ISPRS International Journal of Geo-Information}, 10\penalty0
  (7):\penalty0 485, 2021.

\bibitem[Bai et~al.(2020)Bai, Yao, Li, Wang, and Wang]{bai2020adaptive}
Lei Bai, Lina Yao, Can Li, Xianzhi Wang, and Can Wang.
\newblock Adaptive graph convolutional recurrent network for traffic
  forecasting.
\newblock \emph{Advances in neural information processing systems},
  33:\penalty0 17804--17815, 2020.

\bibitem[Beck et~al.(2017)Beck, Van~Dijk, Levizzani, Schellekens, Miralles,
  Martens, and De~Roo]{beck2017mswep}
Hylke~E Beck, Albert~IJM Van~Dijk, Vincenzo Levizzani, Jaap Schellekens,
  Diego~G Miralles, Brecht Martens, and Ad~De~Roo.
\newblock Mswep: 3-hourly 0.25 global gridded precipitation (1979--2015) by
  merging gauge, satellite, and reanalysis data.
\newblock \emph{Hydrology and Earth System Sciences}, 21\penalty0 (1):\penalty0
  589--615, 2017.

\bibitem[Carrera et~al.(2015)Carrera, B{\'e}lair, and
  Bilodeau]{carrera2015canadian}
Marco~L Carrera, St{\'e}phane B{\'e}lair, and Bernard Bilodeau.
\newblock The canadian land data assimilation system (caldas): Description and
  synthetic evaluation study.
\newblock \emph{Journal of Hydrometeorology}, 16\penalty0 (3):\penalty0
  1293--1314, 2015.

\bibitem[De~Felice et~al.(2013)De~Felice, Alessandri, and
  Ruti]{de2013electricity}
Matteo De~Felice, Andrea Alessandri, and Paolo~M Ruti.
\newblock Electricity demand forecasting over italy: Potential benefits using
  numerical weather prediction models.
\newblock \emph{Electric Power Systems Research}, 104:\penalty0 71--79, 2013.

\bibitem[Defferrard et~al.(2020)Defferrard, Milani, Gusset, and
  Perraudin]{defferrard2020deepsphere}
Micha{\"e}l Defferrard, Martino Milani, Fr{\'e}d{\'e}rick Gusset, and
  Nathana{\"e}l Perraudin.
\newblock Deepsphere: a graph-based spherical cnn.
\newblock \emph{arXiv preprint arXiv:2012.15000}, 2020.

\bibitem[Ebert-Uphoff et~al.(2017)Ebert-Uphoff, Thompson, Demir, Gel, Karpatne,
  Guereque, Kumar, Cabral-Cano, and Smyth]{ebert2017vision}
Imme Ebert-Uphoff, David~R Thompson, Ibrahim Demir, Yulia~R Gel, Anuj Karpatne,
  Mariana Guereque, Vipin Kumar, Enrique Cabral-Cano, and Padhraic Smyth.
\newblock A vision for the development of benchmarks to bridge geoscience and
  data science.
\newblock In \emph{17th International Workshop on Climate Informatics}, 2017.

\bibitem[Geng et~al.(2019)Geng, Li, Wang, Zhang, Yang, Ye, and
  Liu]{geng2019spatiotemporal}
Xu~Geng, Yaguang Li, Leye Wang, Lingyu Zhang, Qiang Yang, Jieping Ye, and Yan
  Liu.
\newblock Spatiotemporal multi-graph convolution network for ride-hailing
  demand forecasting.
\newblock In \emph{Proceedings of the AAAI conference on artificial
  intelligence}, volume~33, pages 3656--3663, 2019.

\bibitem[Guo et~al.(2019)Guo, Lin, Feng, Song, and Wan]{guo2019attention}
Shengnan Guo, Youfang Lin, Ning Feng, Chao Song, and Huaiyu Wan.
\newblock Attention based spatial-temporal graph convolutional networks for
  traffic flow forecasting.
\newblock In \emph{Proceedings of the AAAI conference on artificial
  intelligence}, volume~33, pages 922--929, 2019.

\bibitem[Hammond et~al.(2011)Hammond, Vandergheynst, and
  Gribonval]{hammond2011wavelets}
David~K Hammond, Pierre Vandergheynst, and R{\'e}mi Gribonval.
\newblock Wavelets on graphs via spectral graph theory.
\newblock \emph{Applied and Computational Harmonic Analysis}, 30\penalty0
  (2):\penalty0 129--150, 2011.

\bibitem[He et~al.(2020)He, Yang, Tang, Lu, Qin, Chen, and Li]{he2020first}
Jie He, Kun Yang, Wenjun Tang, Hui Lu, Jun Qin, Yingying Chen, and Xin Li.
\newblock The first high-resolution meteorological forcing dataset for land
  process studies over china.
\newblock \emph{Scientific Data}, 7\penalty0 (1):\penalty0 1--11, 2020.

\bibitem[Hersbach et~al.(2020)Hersbach, Bell, Berrisford, Hirahara,
  Hor{\'a}nyi, Mu{\~n}oz-Sabater, Nicolas, Peubey, Radu, Schepers,
  et~al.]{hersbach2020era5}
Hans Hersbach, Bill Bell, Paul Berrisford, Shoji Hirahara, Andr{\'a}s
  Hor{\'a}nyi, Joaqu{\'\i}n Mu{\~n}oz-Sabater, Julien Nicolas, Carole Peubey,
  Raluca Radu, Dinand Schepers, et~al.
\newblock The era5 global reanalysis.
\newblock \emph{Quarterly Journal of the Royal Meteorological Society},
  146\penalty0 (730):\penalty0 1999--2049, 2020.

\bibitem[Huffman et~al.(2015)Huffman, Bolvin, Braithwaite, Hsu, and
  Joyce]{huffman2015nasa}
George~J Huffman, David~T Bolvin, Dan Braithwaite, Kuolin Hsu, and Robert
  Joyce.
\newblock Nasa global precipitation measurement (gpm) integrated
  multi-satellite retrievals for gpm (imerg).
\newblock 2015.

\bibitem[Kingma and Ba(2014)]{kingma2014adam}
Diederik~P Kingma and Jimmy Ba.
\newblock Adam: A method for stochastic optimization.
\newblock \emph{arXiv preprint arXiv:1412.6980}, 2014.

\bibitem[Kitaev et~al.(2020)Kitaev, Kaiser, and Levskaya]{kitaev2020reformer}
Nikita Kitaev, {\L}ukasz Kaiser, and Anselm Levskaya.
\newblock Reformer: The efficient transformer.
\newblock \emph{arXiv preprint arXiv:2001.04451}, 2020.

\bibitem[Kummerow et~al.(2000)Kummerow, Simpson, Thiele, Barnes, Chang,
  Stocker, Adler, Hou, Kakar, Wentz, et~al.]{kummerow2000status}
Christian Kummerow, J~Simpson, O~Thiele, W~Barnes, ATC Chang, E~Stocker,
  RF~Adler, A~Hou, R~Kakar, F~Wentz, et~al.
\newblock The status of the tropical rainfall measuring mission (trmm) after
  two years in orbit.
\newblock \emph{Journal of applied meteorology}, 39\penalty0 (12):\penalty0
  1965--1982, 2000.

\bibitem[Li et~al.(2019{\natexlab{a}})Li, Han, Cheng, Su, Wang, Zhang, and
  Pan]{li2019predicting}
Jia Li, Zhichao Han, Hong Cheng, Jiao Su, Pengyun Wang, Jianfeng Zhang, and
  Lujia Pan.
\newblock Predicting path failure in time-evolving graphs.
\newblock In \emph{Proceedings of the 25th ACM SIGKDD International Conference
  on Knowledge Discovery \& Data Mining}, pages 1279--1289, 2019{\natexlab{a}}.

\bibitem[Li et~al.(2019{\natexlab{b}})Li, Jin, Xuan, Zhou, Chen, Wang, and
  Yan]{li2019enhancing}
Shiyang Li, Xiaoyong Jin, Yao Xuan, Xiyou Zhou, Wenhu Chen, Yu-Xiang Wang, and
  Xifeng Yan.
\newblock Enhancing the locality and breaking the memory bottleneck of
  transformer on time series forecasting.
\newblock \emph{Advances in neural information processing systems}, 32,
  2019{\natexlab{b}}.

\bibitem[Li et~al.(2017)Li, Yu, Shahabi, and Liu]{li2017diffusion}
Yaguang Li, Rose Yu, Cyrus Shahabi, and Yan Liu.
\newblock Diffusion convolutional recurrent neural network: Data-driven traffic
  forecasting.
\newblock \emph{arXiv preprint arXiv:1707.01926}, 2017.

\bibitem[Lin et~al.(2022)Lin, Gao, Xu, Wu, Li, and Li]{lin2022conditional}
Haitao Lin, Zhangyang Gao, Yongjie Xu, Lirong Wu, Ling Li, and Stan~Z Li.
\newblock Conditional local convolution for spatio-temporal meteorological
  forecasting.
\newblock In \emph{Proceedings of the AAAI Conference on Artificial
  Intelligence}, volume~36, pages 7470--7478, 2022.

\bibitem[Liu et~al.(2021)Liu, Yu, Liao, Li, Lin, Liu, and
  Dustdar]{liu2021pyraformer}
Shizhan Liu, Hang Yu, Cong Liao, Jianguo Li, Weiyao Lin, Alex~X Liu, and
  Schahram Dustdar.
\newblock Pyraformer: Low-complexity pyramidal attention for long-range time
  series modeling and forecasting.
\newblock In \emph{International Conference on Learning Representations}, 2021.

\bibitem[McDonald(2009)]{mcdonald2009ridge}
Gary~C McDonald.
\newblock Ridge regression.
\newblock \emph{Wiley Interdisciplinary Reviews: Computational Statistics},
  1\penalty0 (1):\penalty0 93--100, 2009.

\bibitem[M{\"u}ller and Scheichl(2014)]{muller2014massively}
Eike~H M{\"u}ller and Robert Scheichl.
\newblock Massively parallel solvers for elliptic partial differential
  equations in numerical weather and climate prediction.
\newblock \emph{Quarterly Journal of the Royal Meteorological Society},
  140\penalty0 (685):\penalty0 2608--2624, 2014.

\bibitem[Nielsen et~al.(2021)Nielsen, Iosifidis, and
  Karstoft]{nielsen2021cloudcast}
Andreas~Holm Nielsen, Alexandros Iosifidis, and Henrik Karstoft.
\newblock Cloudcast: A satellite-based dataset and baseline for forecasting
  clouds.
\newblock \emph{IEEE Journal of Selected Topics in Applied Earth Observations
  and Remote Sensing}, 14:\penalty0 3485--3494, 2021.

\bibitem[Panagopoulos et~al.(2021)Panagopoulos, Nikolentzos, and
  Vazirgiannis]{panagopoulos2021transfer}
George Panagopoulos, Giannis Nikolentzos, and Michalis Vazirgiannis.
\newblock Transfer graph neural networks for pandemic forecasting.
\newblock In \emph{Proceedings of the AAAI Conference on Artificial
  Intelligence}, volume~35, pages 4838--4845, 2021.

\bibitem[Paszke et~al.(2019)Paszke, Gross, Massa, Lerer, Bradbury, Chanan,
  Killeen, Lin, Gimelshein, Antiga, et~al.]{paszke2019pytorch}
Adam Paszke, Sam Gross, Francisco Massa, Adam Lerer, James Bradbury, Gregory
  Chanan, Trevor Killeen, Zeming Lin, Natalia Gimelshein, Luca Antiga, et~al.
\newblock Pytorch: An imperative style, high-performance deep learning library.
\newblock \emph{Advances in neural information processing systems}, 32, 2019.

\bibitem[Racah et~al.(2017)Racah, Beckham, Maharaj, Ebrahimi~Kahou, Prabhat,
  and Pal]{racah2017extremeweather}
Evan Racah, Christopher Beckham, Tegan Maharaj, Samira Ebrahimi~Kahou,
  Mr~Prabhat, and Chris Pal.
\newblock Extremeweather: A large-scale climate dataset for semi-supervised
  detection, localization, and understanding of extreme weather events.
\newblock \emph{Advances in neural information processing systems}, 30, 2017.

\bibitem[Rasp et~al.(2020)Rasp, Dueben, Scher, Weyn, Mouatadid, and
  Thuerey]{rasp2020weatherbench}
Stephan Rasp, Peter~D Dueben, Sebastian Scher, Jonathan~A Weyn, Soukayna
  Mouatadid, and Nils Thuerey.
\newblock Weatherbench: a benchmark data set for data-driven weather
  forecasting.
\newblock \emph{Journal of Advances in Modeling Earth Systems}, 12\penalty0
  (11):\penalty0 e2020MS002203, 2020.

\bibitem[Rodell et~al.(2004)Rodell, Houser, Jambor, Gottschalck, Mitchell,
  Meng, Arsenault, Cosgrove, Radakovich, Bosilovich, et~al.]{rodell2004global}
Matthew Rodell, PR~Houser, UEA Jambor, J~Gottschalck, Kieran Mitchell, C-J
  Meng, Kristi Arsenault, B~Cosgrove, J~Radakovich, M~Bosilovich, et~al.
\newblock The global land data assimilation system.
\newblock \emph{Bulletin of the American Meteorological society}, 85\penalty0
  (3):\penalty0 381--394, 2004.

\bibitem[Rozemberczki et~al.(2021)Rozemberczki, Scherer, He, Panagopoulos,
  Riedel, Astefanoaei, Kiss, Beres, L{\'o}pez, Collignon,
  et~al.]{rozemberczki2021pytorch}
Benedek Rozemberczki, Paul Scherer, Yixuan He, George Panagopoulos, Alexander
  Riedel, Maria Astefanoaei, Oliver Kiss, Ferenc Beres, Guzm{\'a}n L{\'o}pez,
  Nicolas Collignon, et~al.
\newblock Pytorch geometric temporal: Spatiotemporal signal processing with
  neural machine learning models.
\newblock In \emph{Proceedings of the 30th ACM International Conference on
  Information \& Knowledge Management}, pages 4564--4573, 2021.

\bibitem[Seo et~al.(2018)Seo, Defferrard, Vandergheynst, and
  Bresson]{seo2018structured}
Youngjoo Seo, Micha{\"e}l Defferrard, Pierre Vandergheynst, and Xavier Bresson.
\newblock Structured sequence modeling with graph convolutional recurrent
  networks.
\newblock In \emph{International conference on neural information processing},
  pages 362--373. Springer, 2018.

\bibitem[Shi et~al.(2019)Shi, Zhang, Cheng, and Lu]{shi2019two}
Lei Shi, Yifan Zhang, Jian Cheng, and Hanqing Lu.
\newblock Two-stream adaptive graph convolutional networks for skeleton-based
  action recognition.
\newblock In \emph{Proceedings of the IEEE/CVF conference on computer vision
  and pattern recognition}, pages 12026--12035, 2019.

\bibitem[Shuman et~al.(2013)Shuman, Narang, Frossard, Ortega, and
  Vandergheynst]{shuman2013emerging}
David~I Shuman, Sunil~K Narang, Pascal Frossard, Antonio Ortega, and Pierre
  Vandergheynst.
\newblock The emerging field of signal processing on graphs: Extending
  high-dimensional data analysis to networks and other irregular domains.
\newblock \emph{IEEE signal processing magazine}, 30\penalty0 (3):\penalty0
  83--98, 2013.

\bibitem[S{\o}nderby et~al.(2020)S{\o}nderby, Espeholt, Heek, Dehghani, Oliver,
  Salimans, Agrawal, Hickey, and Kalchbrenner]{sonderby2020metnet}
Casper~Kaae S{\o}nderby, Lasse Espeholt, Jonathan Heek, Mostafa Dehghani,
  Avital Oliver, Tim Salimans, Shreya Agrawal, Jason Hickey, and Nal
  Kalchbrenner.
\newblock Metnet: A neural weather model for precipitation forecasting.
\newblock \emph{arXiv preprint arXiv:2003.12140}, 2020.

\bibitem[Su et~al.(2012)Su, Yan, and Tsai]{su2012linear}
Xiaogang Su, Xin Yan, and Chih-Ling Tsai.
\newblock Linear regression.
\newblock \emph{Wiley Interdisciplinary Reviews: Computational Statistics},
  4\penalty0 (3):\penalty0 275--294, 2012.

\bibitem[Tolstykh and Frolov(2005)]{tolstykh2005some}
MA~Tolstykh and AV~Frolov.
\newblock Some current problems in numerical weather prediction.
\newblock \emph{Izvestiya Atmospheric and Oceanic Physics}, 41\penalty0
  (3):\penalty0 285--295, 2005.

\bibitem[Vaswani et~al.(2017)Vaswani, Shazeer, Parmar, Uszkoreit, Jones, Gomez,
  Kaiser, and Polosukhin]{vaswani2017attention}
Ashish Vaswani, Noam Shazeer, Niki Parmar, Jakob Uszkoreit, Llion Jones,
  Aidan~N Gomez, {\L}ukasz Kaiser, and Illia Polosukhin.
\newblock Attention is all you need.
\newblock \emph{Advances in neural information processing systems}, 30, 2017.

\bibitem[Vovk(2013)]{vovk2013kernel}
Vladimir Vovk.
\newblock Kernel ridge regression.
\newblock \emph{Empirical Inference: Festschrift in Honor of Vladimir N.
  Vapnik}, pages 105--116, 2013.

\bibitem[Wang et~al.(2020)Wang, Li, Zhang, Meng, Meng, and Gao]{wang2020pm2}
Shuo Wang, Yanran Li, Jiang Zhang, Qingye Meng, Lingwei Meng, and Fei Gao.
\newblock Pm2. 5-gnn: A domain knowledge enhanced graph neural network for pm2.
  5 forecasting.
\newblock In \emph{Proceedings of the 28th International Conference on Advances
  in Geographic Information Systems}, pages 163--166, 2020.

\bibitem[Wu et~al.(2021)Wu, Xu, Wang, and Long]{wu2021autoformer}
Haixu Wu, Jiehui Xu, Jianmin Wang, and Mingsheng Long.
\newblock Autoformer: Decomposition transformers with auto-correlation for
  long-term series forecasting.
\newblock \emph{Advances in Neural Information Processing Systems},
  34:\penalty0 22419--22430, 2021.

\bibitem[Wu et~al.(2020{\natexlab{a}})Wu, Xiao, Ding, Zhao, Wei, and
  Huang]{wu2020adversarial}
Sifan Wu, Xi~Xiao, Qianggang Ding, Peilin Zhao, Ying Wei, and Junzhou Huang.
\newblock Adversarial sparse transformer for time series forecasting.
\newblock \emph{Advances in neural information processing systems},
  33:\penalty0 17105--17115, 2020{\natexlab{a}}.

\bibitem[Wu et~al.(2020{\natexlab{b}})Wu, Pan, Long, Jiang, Chang, and
  Zhang]{wu2020connecting}
Zonghan Wu, Shirui Pan, Guodong Long, Jing Jiang, Xiaojun Chang, and Chengqi
  Zhang.
\newblock Connecting the dots: Multivariate time series forecasting with graph
  neural networks.
\newblock In \emph{Proceedings of the 26th ACM SIGKDD international conference
  on knowledge discovery \& data mining}, pages 753--763, 2020{\natexlab{b}}.

\bibitem[Yu et~al.(2017)Yu, Yin, and Zhu]{yu2017spatio}
Bing Yu, Haoteng Yin, and Zhanxing Zhu.
\newblock Spatio-temporal graph convolutional networks: A deep learning
  framework for traffic forecasting.
\newblock \emph{arXiv preprint arXiv:1709.04875}, 2017.

\bibitem[Zhang et~al.(2020)Zhang, Cao, Zhang, and Xia]{zhang2020crowd}
Xu~Zhang, Ruixu Cao, Zuyu Zhang, and Ying Xia.
\newblock Crowd flow forecasting with multi-graph neural networks.
\newblock In \emph{2020 International Joint Conference on Neural Networks
  (IJCNN)}, pages 1--7. IEEE, 2020.

\bibitem[Zhao et~al.(2019)Zhao, Song, Zhang, Liu, Wang, Lin, Deng, and
  Li]{zhao2019t}
Ling Zhao, Yujiao Song, Chao Zhang, Yu~Liu, Pu~Wang, Tao Lin, Min Deng, and
  Haifeng Li.
\newblock T-gcn: A temporal graph convolutional network for traffic prediction.
\newblock \emph{IEEE Transactions on Intelligent Transportation Systems},
  21\penalty0 (9):\penalty0 3848--3858, 2019.

\bibitem[Zheng et~al.(2020)Zheng, Fan, Wang, and Qi]{zheng2020gman}
Chuanpan Zheng, Xiaoliang Fan, Cheng Wang, and Jianzhong Qi.
\newblock Gman: A graph multi-attention network for traffic prediction.
\newblock In \emph{Proceedings of the AAAI conference on artificial
  intelligence}, volume~34, pages 1234--1241, 2020.

\bibitem[Zhou et~al.(2021)Zhou, Zhang, Peng, Zhang, Li, Xiong, and
  Zhang]{zhou2021informer}
Haoyi Zhou, Shanghang Zhang, Jieqi Peng, Shuai Zhang, Jianxin Li, Hui Xiong,
  and Wancai Zhang.
\newblock Informer: Beyond efficient transformer for long sequence time-series
  forecasting.
\newblock In \emph{Proceedings of the AAAI Conference on Artificial
  Intelligence}, volume~35, pages 11106--11115, 2021.

\bibitem[Zhou et~al.(2022)Zhou, Ma, Wen, Wang, Sun, and Jin]{zhou2022fedformer}
Tian Zhou, Ziqing Ma, Qingsong Wen, Xue Wang, Liang Sun, and Rong Jin.
\newblock Fedformer: Frequency enhanced decomposed transformer for long-term
  series forecasting.
\newblock \emph{arXiv preprint arXiv:2201.12740}, 2022.

\end{thebibliography}

\appendix
\onecolumn

\aistatstitle{Supplementary Materials: Weather2K: A Multivariate Spatio-Temporal Benchmark Dataset for Meteorological Forecasting Based on Real-Time Observation Data from Ground Weather Stations}

\section{NOTATION DEFINITION}

Table~\ref{A} shows the glossary of notations used in this paper.

\begin{table*}[!ht]\small
\renewcommand{\thetable}{5}
\caption{Glossary of notations.} 
\label{A}
\begin{center}
\tabcolsep= 0.7cm
\renewcommand\arraystretch{1.45}
\begin{tabular}{c|l}

\midrule[1pt]

\textbf{Symbol} & \textbf{Meaning}\\ \midrule[1pt]
\multirow{50}{*}
\text{$G =\left(V\text{, } E\text{, } \mathbf{A}  \right) $} & \text{Graph represented nodes, edges and adjacency matrix respectively.} \\ 
\text{${v_i}$} & \text{The $i$-th node.} \\
\text{$N$} & \text{The number of geographic locations.} \\ 
\text{$D$} & \text{The number of used factors.} \\ 
\text{$\mathbf{X}^{(t)}$} & \text{The value of meteorological factors in all locations at the $t$-th interval.} \\ 
\text{$W^{'}$} & \text{The input time length.} \\
\text{$W$} & \text{The forecasting time length.} \\
\text{$F$} & \text{The set of all factors in Weather2K.} \\
\text{$f$} & \text{A specific factor in $F$.} \\
\text{$t_{i, p}^f$} & \text{The time-series data value of the vertex $v_{i}$ at time step $p$ of factor $f$.} \\
\text{$T_i^f$} & \text{The time-series sequence of $v_{i}$ used for training of factor $f$.} \\
\text{$\alpha,\beta$} & \text{A hyper-parameter for controlling the saturation rate of the activation function.} \\
\text{$ \mathbf\Theta$} & \text{The linear layer.} \\
\text{$\mathbf{E}_i$} & \text{The static node embedding of $v_{i}$.} \\
\text{$\mathbf{Z}_i$} & \text{The dynamic node embedding of $v_{i}$.} \\
\text{${S}$} & \text{The set of proposed graphs.} \\
\text{${s}$} & \text{A specific graph in $S$.} \\
\text{${W_s}$} & \text{Learnable weights which describe the importance of graph $s$ for each node.} \\
\text{${\mathbf{Y}^{(t,W)}}$} & \text{The ground truth at time $t$.} \\
\text{$\hat{\mathbf{Y}}^{(t,W)}$} & \text{The final prediction at time $t$.} \\
\text{$\mathbf{x}_t^f$} & \text{The signal of factor $f$ at time $t$ all over the graph.} \\
\text{$g_\theta $} & \text{A kernel to filter the signal.} \\
\text{$ \mathbf L$} & \text{The normalized graph Laplacian.} \\
\text{$\mathbf{U}$} & \text{The matrix of eigenvectors of $ \mathbf L$.} \\
\text{$\mathbf I_N$} & \text{An identity matrix.} \\
\text{$\mathbf D$} & \text{The diagonal degree matrix.} \\
\text{$\mathbf\Lambda$} & \text{The diagonal matrix of eigenvalues of $\mathbf L$} \\
\text{$\tilde{\mathbf{L}}$} & \text{The scaled Laplacian.} \\
\text{$T_k(\tilde{\mathbf{L}})$} & \text{The Chebyshev polynomial of order $k$ evaluated at the scaled Laplacian $\tilde{\mathbf{L}}$.} \\
\midrule[1pt]

\end{tabular}
\end{center}
\end{table*}


\section{METRICS COMPUTATION}
For time series forecasting, let $\mathbf{T}_i \in \mathbf{R}^{L \times C}$ be the ground truth, and $\hat{\mathbf{T}}_i$ be the predictions given by neural networks, where $L$ denotes the forecasting time length and $C$ is the number of forecasting factors. The two metrics including MAE, MSE are calculated as:

\begin{equation}
\renewcommand{\theequation}{14}
\operatorname{MAE}\left(\mathbf{T}_i,\hat{\mathbf{T}}_i\right)=\frac{1}{LC}\left|\hat{\mathbf{T}}_i-\mathbf{T}_i\right|
\end{equation}

\begin{equation}
\renewcommand{\theequation}{15}
\operatorname{MSE}\left(\mathbf{T}_i, \hat{\mathbf{T}}_i \right)= \frac{1}{LC}\left(\hat{\mathbf{T}}_i-\mathbf{T}_i\right)^2
\end{equation}

For spatio-temporal weather forecasting, let $\mathbf{Y}^{(t,W)} = \left[\mathbf{X}^{(t+1)}, \ldots, \mathbf{X}^{(t+W)}\right]$
be the ground truth, and
$\hat{\mathbf{Y}}^{(t,W)} = \left[\hat{\mathbf{X}}^{(t+1)}, \ldots, \hat{\mathbf{X}}^{(t+W)}\right]$
be the predictions given by graph models. The two metrics including MAE, RMSE are calculated as:

\begin{equation}
\renewcommand{\theequation}{16}
\operatorname{MAE}\left(\mathbf{Y}^{(t,W)},\hat{\mathbf{Y}}^{(t,W)}\right)=\frac{1}{WND} \sum_{i=t+1}^{t+W}\left|\hat{\mathbf{X}}^{(i)}-\mathbf{X}^{(i)}\right|
\end{equation}

\begin{equation}
\renewcommand{\theequation}{17}
\operatorname{RMSE}\left(\mathbf{Y}^{(t, W)}, \hat{\mathbf{Y}}^{(t, W)}\right)= \sqrt{\frac{1}{WND}\sum_{i=t+1}^{t+W}\left(\hat{\mathbf{X}}^{(i)}-\mathbf{X}^{(i)}\right)^2}
\end{equation}

where $D,N$ is the number of forecasting factors and geographic locations respectively, $W$ is the forecasting time length.

\section{MORE DETAILS ON THE DATASET OF WEATHER2K}

\subsection{The Data Processing}
For sensor noise, bias and external effects when collecting data, CMA's QC technologies mainly include 
numerical range check, climatic range check, main change range check, internal consistency check, temporal consistency check, spatial consistency check. 
Moreover, there are six main steps in our procedure of data processing, which are listed as follows:

\paragraph{(1) Acquisition of the Raw Data}
The raw data comes from CMA's observation data of China's ground weather stations.
The hourly updated data contains a total of 105 variables from approximately 2,500 ground weather stations covering an area of 6 million square kilometers.
We select the time period from January 2017 to August 2021, in which we can ensure the use of unified meteorological observation instruments and sensors for data collection.

\paragraph{(2) Screening of the Meteorological Factors}
Based on the consideration of building a representative and comprehensive dataset for meteorological forecasting, we select 3 time-invariant constants: latitude, longitude, and altitude to provide position information and 20 important meteorological factors such as air pressure, air temperature, dewpoint temperature, water vapor pressure, relative humidity, wind speed, wind direction, vertical visibility, horizontal visibility, precipitation, etc.
The definitions and physical descriptions of the chosen variables are listed in Table 1.

\paragraph{(3) Screening of the Time Series}
In this step, we mainly ensure the integrity of the time series of each ground weather station. 
According to the chosen time range, the length of the time series should be 40,896 steps for a single ground weather station.
We discard the whole data of a single weather station with a missing proportion above the upper limit, which is set to 1\% of the entire time steps.
For the remaining data, we use linear interpolation.
After this step, the number of eligible ground weather stations drops to 2,373.

\paragraph{(4) Screening of the Default Values}
There are default values for meteorological factors represented by certain numbers, such as 999,999 for visibility.
For a single ground weather station, if the default value proportion for any one of the 20 meteorological factors exceeds 1\%, we will discard the entire data.
We also use linear interpolation for the default values in the remaining data.
After this step, the number of eligible ground weather stations falls further to 2,130.

\paragraph{(5) Manual Tuning of the Outliers}
Despite CMA's QC technologies for the data, we found some unexpected outliers.
We use box plots to make statistics for each meteorological factor.
We focus on discrete individual points in the box plot that are above the upper whisker or below the lower whisker.
The opinions of 3 experts in the field of meteorological science help us decide whether these points are outliers for recording errors or acceptable extremes.

\paragraph{(6) Storage of the Data}
We use a huge Numpy format file to store the processed data.
The historical data of each individual ground weather station constitutes the time series of multivariate meteorological factors.
Thousands of stations form the spatial sequences, each of which is assigned a fixed index number based on the original station number.
In the release version of the dataset, we give the index correspondence in a text file.

\subsection{Dataset Statistics}

\subsubsection{Statistics of Meteorological Factors by Box Plot}

\begin{figure}[!ht] 
\renewcommand{\thefigure}{4}
\centerline{\includegraphics[scale=0.49]{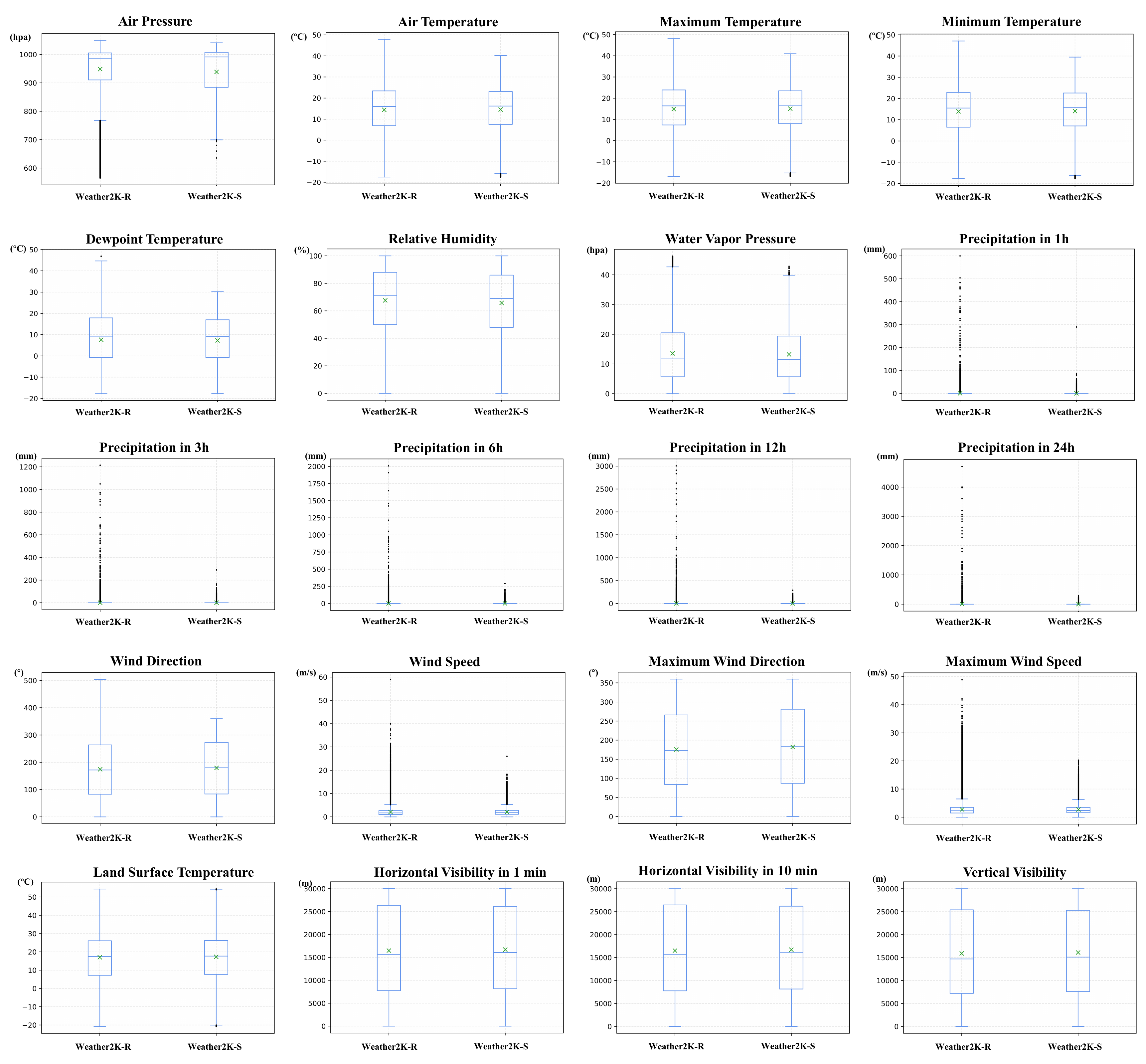}}
\vspace{.1in}
\caption{Box plots of 20 meteorological factors in Weather2K-R and Weather2K-S.}
\label{C.2.1}
\end{figure}


In descriptive statistics, the box plot is a method for graphically reflecting the overall characteristics of numerical data distribution through their quartiles.
Although we cannot apply a simple Gaussian distribution for modelling various weather properties, such as precipitation and wind, it can also play an auxiliary role in outlier processing.
Figure~\ref{C.2.1} shows the box plots of 20 meteorological factors in Weather2K-R and Weather2K-S.
The top and bottom boundaries of a box are the upper and lower quartiles of the statistic indices at these stations, while the line and the cross inside the box are the median and the mean value, respectively.
The vertical dashed lines extending from the box represent the minimum and maximum of the corresponding indices. 
"Outliers" (acceptable after expert discussion) are plotted as individual black points.

\subsubsection{Statistics of Weather2K-S by Cumulative Distribution Function}

\begin{figure}[!ht] 
\renewcommand{\thefigure}{5}
\centerline{\includegraphics[scale=0.47]{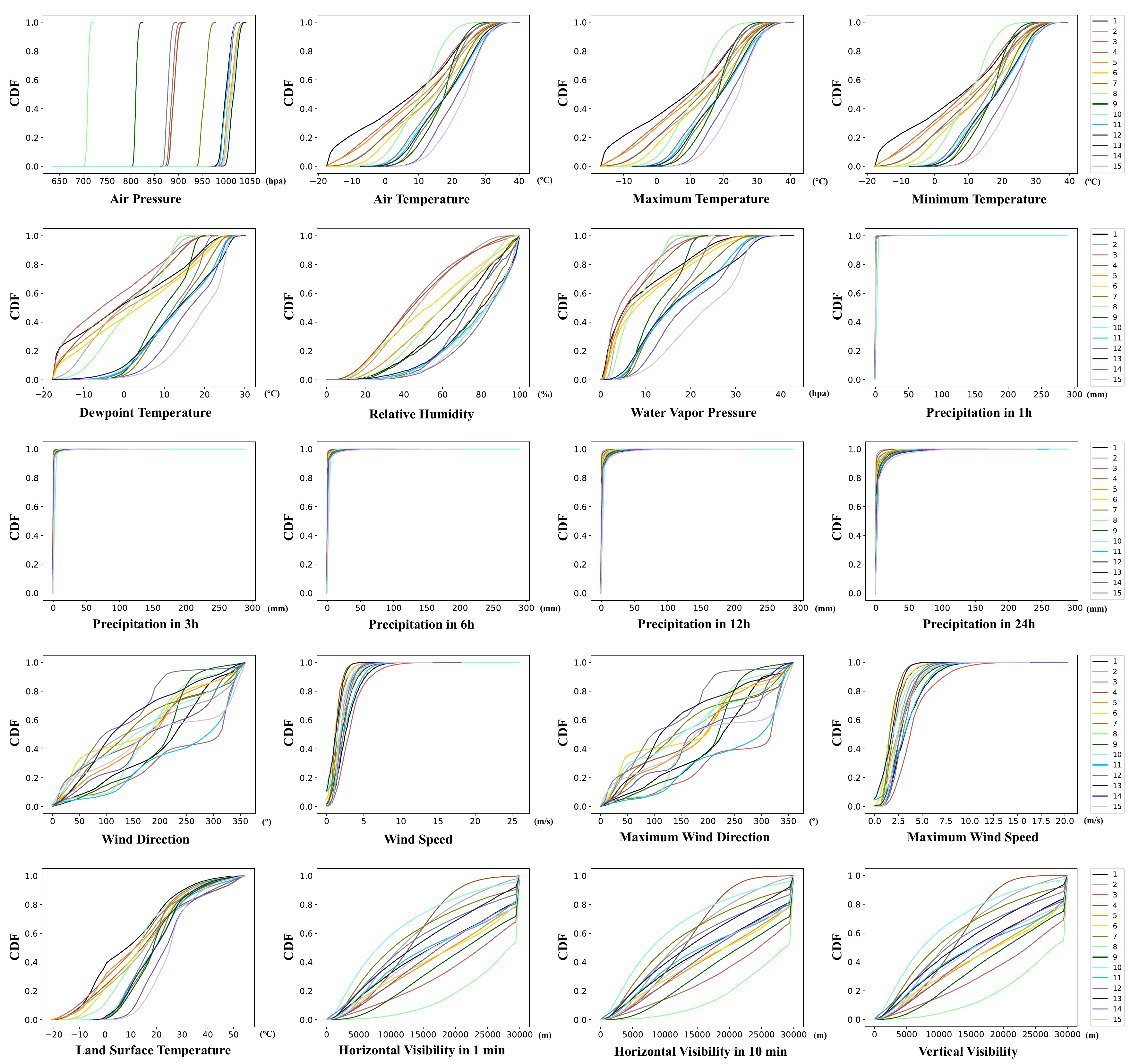}}
\vspace{.1in}
\caption{CDF of 20 meteorological factors in Weather2K-S.}
\label{C.2.2}
\end{figure}

In probability theory and statistics, the Cumulative Distribution Function (CDF) can completely describes the probability distribution of a real-valued variable.
Figure~\ref{C.2.2} shows the CDF of 20 meteorological factors in Weather2K-S.
Lines 1-15 denote the different ground weather stations in Weather2K-S, which can be intuitively found that the stations we select are representative and diverse.

\subsubsection{Statistics of Weather2K-R by Administrative Region}

\begin{figure}[!ht] 
\renewcommand{\thefigure}{6}
\centerline{\includegraphics[scale=0.55]{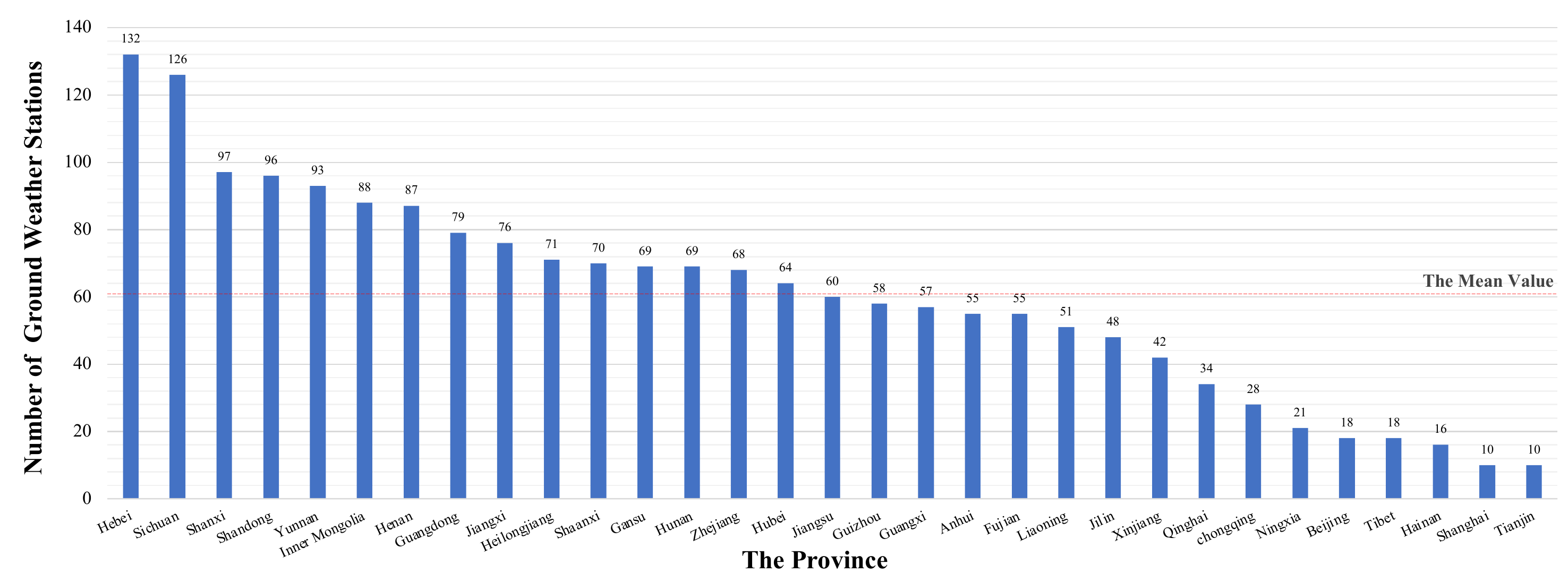}}
\vspace{.1in}
\caption{The statistics of ground weather stations of each province in Weather2K-R.}
\label{C.2.4}
\end{figure}

Figure~\ref{C.2.4} shows the statistics of ground weather stations of each province in Weather2K-R to facilitate the potential use of specific audiences.
Weather2K-R covers 31 provinces in China and contains 1,866 ground weather stations, with an average of more than 60 ground weather stations per province.
Among them, Hebei has the largest number of 132 stations, while Shanghai and Tianjin have the smallest number of 10 stations.


\section{EXPERIMENT}

\subsection{Baseline Models}

In the application direction of time series forecasting, four representative transformer-based baselines are set up for comparison on Weather2K-S.

\textbf{Transformer  }
\citet{vaswani2017attention} propose a simple network architecture, the Transformer, based on the self-attention mechanism.
Transformers show great modeling ability for long-range dependencies and interactions in sequential data and thus are attractive in time series modeling.
However, it is computationally prohibitive to applying self-attention to long-term time series forecasting due to the quadratic complexity of sequence length $L$ in both memory and time.
Many Transformer variants have been proposed to adapt to long-term time series forecasting in a efficient way.

\textbf{Reformer  }
\citet{kitaev2020reformer} introduce two techniques to improve the efficiency of Transformers.
For one, Reformer replaces dot-product attention by one that uses locality-sensitive hashing, changing its complexity from $\mathcal{O}(L^{2})$ to $\mathcal{O}(L \log L)$.
Furthermore, Reformer uses reversible residual layers instead of the standard residuals, which allows storing activations only once in the training process instead of N times, where N is the number of layers. 

\textbf{Informer  }
\citet{zhou2021informer} propose ProbSparse self-attention mechanism to efficiently replace the canonical self-attention, which use the Kullback-Leibler divergence distribution measurement between queries and keys to select $\mathcal{O}(\log L)$ dominant queries.
Informer achieves the $\mathcal{O}(L \log L)$ time complexity and memory usage on dependency alignments.
It also designs a generative style decoder to produce long sequence output with only one forward step to avoid accumulation error.

\textbf{Autoformer  }
\citet{wu2021autoformer} devise a seasonal-trend decomposition architecture with an auto-correlation mechanism working as an attention module.
The auto-correlation block measures the time-delay similarity between inputs signal and aggregate the top-k similar sub-series to produce the output.
Autoformer achieves $\mathcal{O}(L \log L)$ complexity and breaks the information utilization bottleneck by expanding the point-wise representation aggregation to sub-series level.

In the application direction of spatio-temporal forecasting, eight state-of-the-art spatio-temporal GNN models are set up for comparison on Weather2K-R.

\textbf{STGCN  }
\citet{yu2017spatio} propose a deep learning framework for traffic prediction, integrating graph convolution and gated temporal convolution through spatio-temporal convolutional blocks.
The model is entirely composed of convolutional structures and therefore achieves parallelization over input with fewer parameters and faster training speed.

\textbf{DCRNN  }
\citet{li2017diffusion} study the traffic forecasting problem and model the spatial dependency of traffic as a diffusion process on a directed graph.
DCRNN captures both spatial and temporal dependencies among time series using diffusion convolution and the sequence to sequence learning framework together with scheduled sampling. 

\textbf{GCGRU  }
\citet{seo2018structured} combine convolutional neural networks on graphs to identify spatial structures and recurrent neural networks to find dynamic patterns. GCGRU shows that exploiting simultaneously graph spatial and dynamic information about data can improve both precision and learning speed.

\textbf{ASTGCN  }
\citet{guo2019attention} develop a attention mechanism to learn the dynamic spatio-temporal correlations of traffic data. 
The spatio-temporal convolution module consists of graph convolutions for capturing spatial features from the traffic network structure and convolutions in the temporal dimension for describing dependencies from nearby time slices.

\textbf{MSTGCN  }
\citet{guo2019attention} propose a degraded version of ASTGCN, which gets rid of the spatio-temporal attention.

\textbf{TGCN  }
\citet{zhao2019t} propose the TGCN model by combining the graph convolutional network and gated recurrent unit, which are used to capture the topological structure of the road network to model spatial dependence and the dynamic change of traffic data on the roads to model temporal dependence, respectively.

\textbf{A3TGCN  }
\citet{bai2021a3t} propose the A3TGCN model to simultaneously capture global temporal dynamics and spatial correlations. 
The attention mechanism was introduced to adjust the importance of different time points and assemble global temporal information to improve prediction accuracy.

\textbf{CLCRN  }
Based on the assumption of smoothness of location-characterized patterns, \citet{lin2022conditional} propose the conditional local kernel and embed it in a graph-convolution-based recurrent network to model the temporal dynamics for spatio-temporal meteorological forecasting.

\subsection{Implementation Details}
All the deep learning networks mentioned in this paper are implemented in PyTorch \citep{paszke2019pytorch} of version 1.8.0 with CUDA version 11.1.
For the hyper-parameters that are not specifically mentioned in each model, we use the default settings from their original proposals.
Every single experiment is executed on a server with 4 NVIDIA GeForce RTX3090 GPUs with 24GB of video memory.

\subsubsection{Transformers in Time Series Forecasting}

Experiments are carried out on Weather2K-S with the temporal scale from January 1, 2019 to December 31, 2020.
The data is divided into the training set, validation set, and test set according to the ratio of 3:1:2. 
The batch size is set to 32.
The global seed is set to 2021 for the experiment repeat.
The train epoch is set to 30, while the training process is early stopped within 5 epochs if no loss degradation on the validation set is observed.
All the models are trained with L2 loss, using the Adam \citep{kingma2014adam} optimizer with the learning rate of 1$e^{-4}$.

\subsubsection{GNNs in Spatio-Temporal Forecasting}

We truncate Weather2K-R with the temporal scale from January 1, 2017 to December 31, 2019.
The baselines are all implemented according to PyTorch Geometric Temporal \citep{rozemberczki2021pytorch}.
We construct a distance graph as one of the GNN models' inputs by utilizing three time-invariant constants in Weather2K-R that provide position information including latitude, longitude, and altitude.
The data for training, validation, and testing are all non-overlapping one-year time, while the random seed is set to 2022.
The batch size for training is set to 32.
All the models are trained with the target function of MAE and optimized by Adam optimizer \citep{kingma2014adam} for 100 epochs.
Early-stopping epoch is set to 50 according to the validation set. 
The initial learning rate is set to 1$e^{-2}$, and it decays with the ratio 5$e^{-2}$ per 10 epochs in the first 50 epochs.

\subsection{The Results of Multivariate to Univariate Forecasting with Transformer Baselines}
To take full advantage of multivariate data in Weather2K-S, we explore and attempt to use the multivariate data to forecast the target univariate meteorological factor. 
We conduct experiments on the assumption that there are some underlying correlations between different meteorological factors.
Without any prior knowledge, we use all 20 dimensional meteorological factors as input data.
Table~\ref{D3} shows the performance of multivariate to univariate forecasting.
The number in the parenthesis is the difference value from the univariate result, where the better result is highlighted in bold.
However, multivariate to univariate forecasting does not result in stable performance gain and even brought performance degradation in most cases compared to univariate forecasting.
It can be concluded that multivariate information of meteorological factors is underutilized in existing transformer-based time series forecasting baselines.

\begin{table*}[!ht]
\tiny
\renewcommand{\thetable}{6}
\caption{Multivariate to univariate results with different prediction lengths of 24, 72, 168, 336, and 720 on Weather2K-S. The input time length is 72 steps. The better results compared to univariate results are highlighted in \textbf{bold}.} \label{D3}
\begin{center}
\tabcolsep= 0.2cm
\renewcommand\arraystretch{1.1}
\begin{tabular}{c|c|c c|c c|c c|c c }
\midrule[1pt]
\multicolumn{2}{c|}{\textbf{Models}} & \multicolumn{2}{c|}{Transformer}& \multicolumn{2}{c|}{Reformer} & \multicolumn{2}{c|}{Informer} & \multicolumn{2}{c}{Autoformer} \\ \midrule[1pt]

\textbf{Factors} & \textbf{Metrics} & MSE & MAE & MSE & MAE & MSE & MAE & MSE & MAE  \\ \midrule[1pt]

\multirow{5}{*}{\rotatebox{90}{Temperature}} & 24 & 0.0874 (+0.0134) & 0.2256 (+0.0256) & 0.0820 \textbf{(-0.0077)} & 0.2173 \textbf{(-0.0133)} & 0.0953 (+0.0195) & 0.2322 (+0.0291) & 0.1114 (+0.0136) & 0.2539 (+0.0172) \\ 
& 72 & 0.1379 (+0.0133) & 0.2843 (+0.0187) & 0.1200 \textbf{(-0.0094)} & 0.2669 \textbf{(-0.0128)} & 0.1575 (+0.0303) & 0.3033 (+0.0322) & 0.1505 (+0.0121) & 0.2968 (+0.0136)  \\ 
& 168 & 0.1624 (+0.0064) & 0.3090 (+0.0091) & 0.1354 \textbf{(-0.0077)} & 0.2841 \textbf{(-0.0085)} & 0.2002 (+0.0303) & 0.3450 (+0.0288) & 0.1642 (+0.0068) & 0.3078 (+0.0076) \\ 
& 336 & 0.1757 (+0.0048) & 0.3224 (+0.0076) & 0.1398 \textbf{(-0.0053)} & 0.2886 \textbf{(-0.0061)} & 0.2298 (+0.0180) & 0.3785 (+0.0199) & 0.1734 (+0.0079) & 0.3180 (+0.0088) \\
& 720& 0.1939 (+0.0336) & 0.3399 (+0.0341) & 0.1468 \textbf{(-0.0013)} & 0.2940 \textbf{(-0.0028)} & 0.2550 (+0.0431) & 0.3978 (+0.0341) & 0.2037 (+0.0026) & 0.3486 (+0.0033) \\\midrule[1pt]

\multirow{5}{*}{\rotatebox{90}{Visibility}} & 24 & 0.7300 (+0.0195) & 0.6696 (+0.0160) & 0.6355 \textbf{(-0.0268)} & 0.6101 \textbf{(-0.0187)} & 0.7316 (+0.0309) & 0.6555 (+0.0165) & 0.8596 (+0.0582) & 0.7188 (+0.0259) \\ 
& 72 & 0.8943 (+0.0221) & 0.7548 (+0.0073) & 0.8072 \textbf{(-0.0232)} & 0.7219 \textbf{(-0.0221)} & 0.9388 (+0.0541) & 0.7587 (+0.0084) & 1.0087 (+0.0737) & 0.7838 (+0.0253)  \\ 
& 168 & 0.9101 \textbf{(-0.0052)} & 0.7721 \textbf{(-0.0081)} & 0.9067 \textbf{(-0.0016)} & 0.7672 \textbf{(-0.0123)} & 0.9853 (+0.0425) & 0.8014 (+0.0071) & 1.0108 (+0.0197) & 0.7873 (+0.0039) \\
& 336 & 0.9883 \textbf{(-0.0069)} & 0.7989 \textbf{(-0.0070)} & 0.9203 \textbf{(-0.0016)} & 0.7787 \textbf{(-0.0023)} & 1.0229 \textbf{(-0.0003)} & 0.8304 \textbf{(-0.0084)} & 1.0135 (+0.0216) & 0.7907 (+0.0075)  \\
& 720 & 1.0143 (+0.0180) & 0.8117 (+0.0027) & 0.9453 \textbf{(-0.0053)} & 0.7964 (+0.0018) & 1.0736 (+0.0501) & 0.8539 (+0.0101) & 1.0596 \textbf{(-0.0055)} & 0.8092 \textbf{(-0.0053)} \\\midrule[1pt]

\multirow{5}{*}{\rotatebox{90}{Humidity}} & 24 & 0.4307 (+0.0561) & 0.5010 (+0.0403) & 0.3632 (+0.0014) & 0.4551 (+0.0022) & 0.4318 (+0.0569) & 0.5024 (+0.0461) & 0.5217 (+0.0926) & 0.5539 (+0.0598) \\ 
& 72 & 0.5741 (+0.0513) & 0.5863 (+0.0260) & 0.4941 (+0.0009) & 0.5454 \textbf{(-0.0001}) & 0.5989 (+0.0757) & 0.5985 (+0.0366) & 0.6215 (+0.0704) & 0.6140 (+0.0419)   \\ 
& 168 & 0.6427 (+0.0492) & 0.6242 (+0.0238) & 0.5457 (+0.0036) & 0.5767 (+0.0008) & 0.7281 (+0.1060) & 0.6617 (+0.0454) & 0.6536 (+0.0425) & 0.6283 (+0.0241) \\ 
& 336 & 0.6620 (+0.0513) & 0.6354 (+0.0212) & 0.5856 (+0.0095) & 0.5992 (+0.0054) & 0.8020 (+0.1055) & 0.7129 (+0.0476) & 0.6896 (+0.0382) & 0.6456 (+0.0198) \\
& 720& 0.6668 (+0.0491) & 0.6359 (+0.0213) & 0.5819 \textbf{(-0.0045)} & 0.5963 \textbf{(-0.0046)} & 0.8803 (+0.0824) & 0.7711 (+0.0426) & 0.7358 (+0.0213) & 0.6719 (+0.0120) \\\midrule[1pt]

\end{tabular}
\end{center}
\end{table*}

\subsection{The Univariate Results of the Persistent Model and Classical Nonparametric Methods}

In order to perform a meaningful comparison for the forecasting results, 
the persistent model should be introduced to quantify the improvement given by advanced forecasting techniques.
The persistent model is the most cost-effective forecasting model \citep{de2013electricity} which assumes that the conditions will not change, which means “today's weather is tomorrow's forecast”.
So it is often called the naive forecast.
In addition, we present the univariate forecasting results of 3 classical nonparametric methods including linear regression \citep{su2012linear}, ridge regression \citep{mcdonald2009ridge}, and kernel ridge regression \citep{vovk2013kernel} on Weather2K-S in Table~\ref{D4}.
Different from transformer baselines, we use all the data of Weather2K-S, with a 3:1 ratio of the training set to the test set.

\begin{table*}[!ht]
\tiny
\renewcommand{\thetable}{7}
\caption{Univariate results of the persistent model and nonparametric methods with different prediction lengths of 24, 72, 168, 336, and 720 steps. The input time length is 72 steps. The reported results of mean and standard deviation are obtained through experiments on Weather2K-S. Results with \textbf{bold} and \underline{underlines} are the best and worst performance, respectively.} \label{D4}

\begin{center}
\tabcolsep= 0.22cm
\renewcommand\arraystretch{1.1}
\begin{tabular}{c|c|c c|c c|c c|c c }
\midrule[1pt]
\multicolumn{2}{c|}{\textbf{Models}} & \multicolumn{2}{c|}{Persistent Model}& \multicolumn{2}{c|}{Linear Regression} & \multicolumn{2}{c|}{Ridge Regression} & \multicolumn{2}{c}{Kernel Ridge Regression} \\ \midrule[1pt]

\textbf{Factors} & \textbf{Metrics} & MSE & MAE & MSE & MAE & MSE & MAE & MSE & MAE  \\ \midrule[1pt]

\multirow{5}{*}{Temperature} & 24 & \underline{0.2610}$\pm$0.1253 & \underline{0.3771}$\pm$0.0883 & \textbf{0.0712}$\pm$0.0222 & \textbf{0.1892}$\pm$0.0269 & 0.0713$\pm$0.0222 & 0.1893$\pm$0.0269 & 0.0713$\pm$0.0222 & 0.1893$\pm$0.0269 \\ 
& 72 & \underline{0.3219}$\pm$0.1336 & \underline{0.4285}$\pm$0.0855 & \textbf{0.1237}$\pm$0.0417 & \textbf{0.2548}$\pm$0.0389 & 0.1237$\pm$0.0418 & 0.2549$\pm$0.0389 & 0.1237$\pm$0.0418 & 0.2550$\pm$0.0390 \\ 
& 168 & \underline{0.3793}$\pm$0.1501 & \underline{0.4697}$\pm$0.0892 & \textbf{0.1653}$\pm$0.0608 & \textbf{0.3003}$\pm$0.0510 & 0.1655$\pm$0.0608 & 0.3005$\pm$0.0510 & 0.1654$\pm$0.0609 & 0.3006$\pm$0.0514 \\ 
& 336 & \underline{0.4179}$\pm$0.1595 & \underline{0.4968}$\pm$0.0909 & \textbf{0.1962}$\pm$0.0689 & \textbf{0.3329}$\pm$0.0538 & 0.1963$\pm$0.0689 & 0.3331$\pm$0.0538 & 0.1964$\pm$0.0691 & 0.3332$\pm$0.0544 \\
& 720 & \underline{0.4951}$\pm$0.1598 & \underline{0.5463}$\pm$0.0847 & \textbf{0.2632}$\pm$0.0696 & \textbf{0.3931}$\pm$0.0472 & 0.2634$\pm$0.0696 & 0.3932$\pm$0.0472 & 0.2635$\pm$0.0701 & 0.3943$\pm$0.0478 \\\midrule[1pt]

\multirow{5}{*}{Visibility} & 24 & \underline{1.0079}$\pm$0.2200 & \underline{0.6662}$\pm$0.1353 & 0.6355$\pm$0.1253 & \textbf{0.6147}$\pm$0.1007 & \textbf{0.6350}$\pm$0.1261 & 0.6162$\pm$0.1009 & 0.6357$\pm$0.1255 & 0.6148$\pm$0.1009 \\ 
& 72 & \underline{1.3548}$\pm$0.3239 & \underline{0.8147}$\pm$0.1712 & 0.8115$\pm$0.1815 & \textbf{0.7355}$\pm$0.1210 & \textbf{0.8114}$\pm$0.1821 & 0.7366$\pm$0.1214 & 0.8121$\pm$0.1820 & 0.7357$\pm$0.1213  \\ 
& 168 & \underline{1.5628}$\pm$0.3818 & \underline{0.8998}$\pm$0.1939 & 0.9038$\pm$0.2073 & \textbf{0.7948}$\pm$0.1260 & \textbf{0.9037}$\pm$0.2077 & 0.7954$\pm$0.1262 & 0.9047$\pm$0.2083 & 0.7950$\pm$0.1265  \\ 
& 336 & \underline{1.6512}$\pm$0.4031 & \underline{0.9364}$\pm$0.2027 & \textbf{0.9434}$\pm$0.2298 & \textbf{0.8207}$\pm$0.1317 & 0.9447$\pm$0.2312 & 0.8233$\pm$0.1323 & 0.9446$\pm$0.2308 & 0.8210$\pm$0.1322  \\
& 720 & \underline{1.7290}$\pm$0.4237 & \underline{0.9665}$\pm$0.2113 & \textbf{0.9801}$\pm$0.2481 & \textbf{0.8417}$\pm$0.1366 & 0.9812$\pm$0.2488 & 0.8444$\pm$0.1367 & 0.9810$\pm$0.2485 & 0.8418$\pm$0.1369 \\ \midrule[1pt]

\multirow{5}{*}{Humidity} & 24 & \underline{1.0022}$\pm$0.1534 & \underline{0.7505}$\pm$0.0610 & 0.3548$\pm$0.0796 & 0.4378$\pm$0.0529 & \textbf{0.3547}$\pm$0.0796 & 0.4380$\pm$0.0529 & \textbf{0.3547}$\pm$0.0796 & \textbf{0.4378}$\pm$0.0528 \\ 
& 72 & \underline{1.2302}$\pm$0.1741 & \underline{0.8504}$\pm$0.0608 & \textbf{0.4939}$\pm$0.1060 & \textbf{0.5356}$\pm$0.0625 & 0.4940$\pm$0.1060 & 0.5357$\pm$0.0625 & \textbf{0.4939}$\pm$0.1060 & \textbf{0.5356}$\pm$0.0625  \\ 
& 168 & \underline{1.3995}$\pm$0.2021 & \underline{0.9172}$\pm$0.0619 & 0.5770$\pm$0.1179 & 0.5909$\pm$0.0647 & 0.5776$\pm$0.1180 & 0.5918$\pm$0.0648 & \textbf{0.5769}$\pm$0.1178 & \textbf{0.5908}$\pm$0.0646 \\ 
& 336 & \underline{1.5181}$\pm$0.2296 & \underline{0.9608}$\pm$0.0656 & 0.6281$\pm$0.1240 & 0.6233$\pm$0.0663 & 0.6285$\pm$0.1240 & 0.6239$\pm$0.0664 & \textbf{0.6278}$\pm$0.1239 & \textbf{0.6231}$\pm$0.0662  \\
& 720& \underline{1.6328}$\pm$0.2538 & \underline{1.0030}$\pm$0.0707 & 0.6723$\pm$0.1316 & 0.6500$\pm$0.0693 & 0.6726$\pm$0.1316 & 0.6505$\pm$0.0693 & \textbf{0.6718}$\pm$0.1316 & \textbf{0.6496}$\pm$0.0691 \\\midrule[1pt]

\end{tabular}
\end{center}
\end{table*}


\subsection{Discussion of the value of $N_A$ in Neighbor Graph}

\begin{figure}[!ht] 
\renewcommand{\thefigure}{7}
\centerline{\includegraphics[scale=0.43]{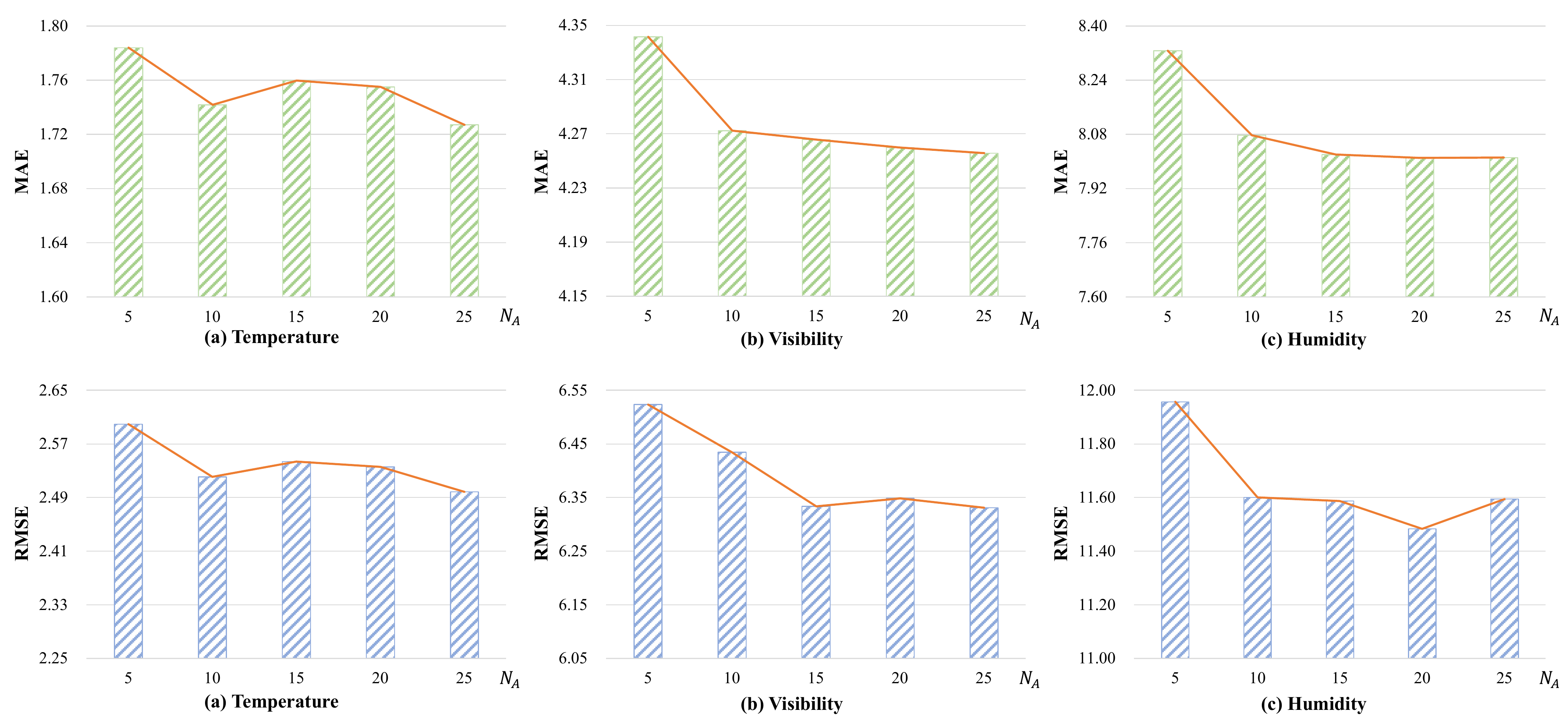}}
\vspace{.1in}
\caption{The selection of $N_A$ in neighbor graph in forecasting (a) temperature, (b) visibility, and (c) humidity.}
\label{D.5}
\end{figure}

When we use the neighbor graph $G_N$ separately, the influence of the selection value of $N_A$ should be discussed.
In Figure~\ref{D.5}, we respectively set $N_A$ as 5, 10, 15, 20, and 25 to verify the variation trend of MAE and RMSE when the forecasting time step is 12.
The performance will not be significantly improved when the number exceeds 10.
Therefore, we consistently set $N_A$ to 10 in the following ablation studies.

\vfill

\end{document}